\documentclass{article}

\usepackage{iclr2026_conference,times}

\usepackage{amsmath,amsfonts,bm}

\def\eqref#1{equation~\ref{#1}}

\def\1{\bm{1}}

\DeclareMathAlphabet{\mathsfit}{\encodingdefault}{\sfdefault}{m}{sl}
\SetMathAlphabet{\mathsfit}{bold}{\encodingdefault}{\sfdefault}{bx}{n}

\usepackage[utf8]{inputenc} 
\usepackage[T1]{fontenc}    
\usepackage{hyperref}       
\usepackage{url}            
\usepackage{booktabs}
\usepackage{amsfonts}       
\usepackage{nicefrac}       
\usepackage{microtype}      
\usepackage{xcolor}         
\usepackage{amsmath}
\usepackage{xspace}

\usepackage{booktabs}       
\usepackage{graphicx}
\usepackage{multirow}
\usepackage{multicol}
\usepackage{color}
\usepackage{colortbl}
\definecolor{Gray}{gray}{0.86}
\usepackage[most]{tcolorbox}
\usepackage{tabularx}

\usepackage{amssymb}
\usepackage{bbding}
\usepackage{pifont}
\usepackage{wasysym}

\usepackage{algorithm}
\usepackage{algpseudocode}
\usepackage{amsmath}
\usepackage{caption}
\usepackage{float}
\usepackage{subcaption}
\usepackage{wrapfig}

\title{ReDDiT: Rehashing Noise for Discrete \\Visual Generation}
\author{%
    \begin{minipage}[t]{\textwidth}
        \textbf{Tianren Ma}$^1$ \ \ \textbf{Xiaosong Zhang}$^1$ \ \ \textbf{Boyu Yang}$^2$ \ \ \textbf{Junlan Feng}$^2$ \ \ \textbf{Qixiang Ye}$^1$ 
    \end{minipage}
}
\iclrfinalcopy
\begin{document}
\maketitle
\vspace{-1cm}
\ \ $^1$ University of Chinese Academy of Sciences  
$^2$ China Mobile Research Institute

\vspace{-0.2cm}
\texttt{\small \ \ matianren18@mails.ucas.ac.cn \ \ qxye@ucas.ac.cn}

\begin{abstract}
In the visual generative area, discrete diffusion models are gaining traction for their efficiency and compatibility. 
However, pioneered attempts still fall behind their continuous counterparts, which we attribute to noise (absorbing state) design and sampling heuristics.
In this study, we propose a rehashing noise approach for discrete diffusion transformer (termed \textbf{ReDDiT}), 
with the aim to extend absorbing states and improve expressive capacity of discrete diffusion models. 
ReDDiT enriches the potential paths that latent variables traverse during training with randomized multi-index corruption. 
The derived rehash sampler, which reverses the randomized absorbing paths, guarantees high diversity and low discrepancy of the generation process.
These reformulations lead to more consistent and competitive generation quality, mitigating the need for heavily tuned randomness.
Experiments show that ReDDiT significantly outperforms the baseline model (reducing gFID from 6.18 to \textbf{1.61}) and is on par with the continuous counterparts. 
The code and models will be publicly available.
\end{abstract}

\section{Introduction}
\label{sec:intro}

Diffusion has been a competitive approach for generative workloads~\cite{dhariwalDiffusionModelsBeat2021,rombachHighResolutionImageSynthesis2022a, liHunyuanDiTPowerfulMultiResolution2024}, offering strong bidirectional perception and well-structured mechanisms~\cite{zhangAddingConditionalControl2023} for global control over content.
Within the continuous domain, diffusion transformers (DiTs)~\cite{peeblesScalableDiffusionModels2023}, which progressively refine image latents from Gaussian noise, have achieved impressive and scalable results.
Recently, the community shows a growing interest in discrete diffusion models~\cite{ huMASKAllYou2024, swerdlowUnifiedMultimodalDiscrete2025},
which is based on their practical advantages, \textit{e.g.}, compatibility with language models for the indexable codebook, and efficiency for predicting multiple tokens at each inference. 
%
Early endeavors~\cite{changMaskGITMaskedGenerative2022, changMuseTextImageGeneration2023, guVectorQuantizedDiffusion2022} pursue efficiency through integrating visual tokenizers and BERT-style \texttt{[mask]} tokens~\cite{devlinBERTPretrainingDeep2019}.
Recent studies~\cite{baiMeissonicRevitalizingMasked2025,mmada2025} improved the generation quality, demonstrating great potential of discrete diffusion.

\begin{figure}[t]
\centering
\includegraphics[width=\linewidth]{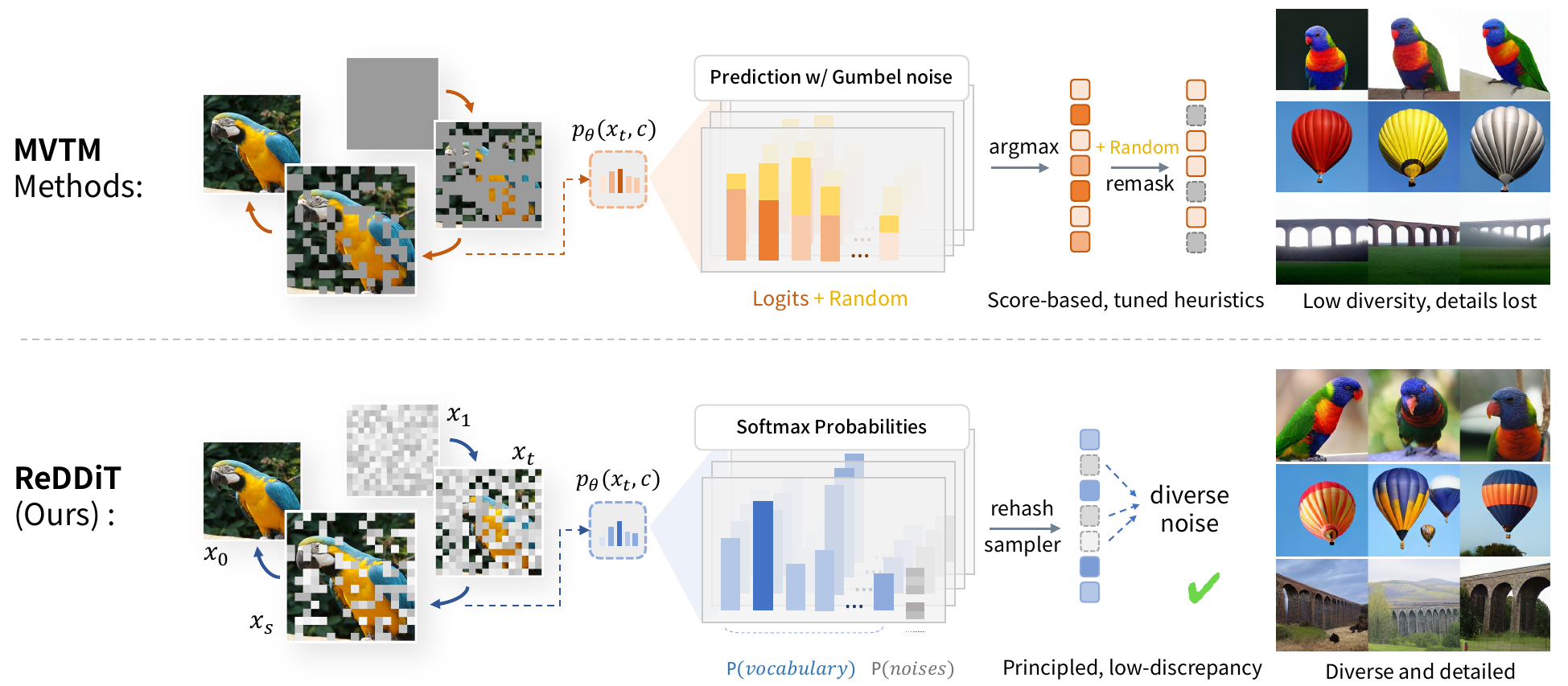}
\caption{\textbf{Comparison the baseline discrete model (MVTM) with ReDDiT.} MVTMs rely on score-based remasking strategies with Gumbel-max to sample from logits, which leads to lower token diversity and suboptimal token selection.
In contrast, ReDDiT introduces a systematic,
low-discrepancy rehashing mechanism that leverages softmax-based probabilities, enabling diverse, high-quality sampling through a learned distribution. (This figure is best viewed in color)}
\label{fig1}
\end{figure}

Despite the progress, the performance of discrete diffusion methods remains lagging behind their continuous counterparts.
%
Representative approaches, \textit{e.g.}, masked visual token models (MVTMs)~\cite{changMaskGITMaskedGenerative2022, yuLanguageModelBeats2024}, are puzzled by the mask design and confidence-based re-mask sampler, 
which restricts model's expressive capacity and makes prediction sensitive to adaptions given extensive training, Fig.~\ref{fig1}(upper).
Moreover, when paired with large-vocabulary codebooks from high-fidelity modern tokenizers, they encounter challenges such as slower sampling speeds and numerical inaccuracy~\cite{zhengMaskedDiffusionModels2025}.

To address these limitations, we first propose two hypotheses.
First, while discrete methods learn to recover plausible tokens from a monotonous \texttt{[mask]} canvas, the used noise design may not be well-suited for discrete visual generation.
In continuous diffusion, Gaussian noise is used to progressively degrade the input to learn a smooth distribution shift ~\cite{hoDenoisingDiffusionProbabilistic2020,luDPMSolverFastSolver2023}.
Discrete masking mimics this paradigm by collapsing all masked tokens to a single absorbing state, which, however, lacks the variability of Gaussian noise, in terms of both vocabulary richness and latent diversity.
Consequently, the discrete process offers a far coarser signal, which limits its ability to represent diverse data distributions~\cite{santosBlackoutDiffusionGenerative2023,austinStructuredDenoisingDiffusion2023}. 
Moreover, while continuous diffusion models introduce stochasticity at every inference step through noise injection, discrete unmasking is inherently binary: tokens are either masked or deterministically decoded, Fig.~\ref{fig1}(upper). This rigid mechanism constrains the flexibility of sample refinement during generation.

Second, the confidence-based re-mask sampler of MVTMs introduces a form of handcrafted randomness, which is implemented through Gumbel-max, to approximate sampling diversity.
Unfortunately, this sampler compromises the probabilistic fidelity of generation, and the need to carefully balance token numbers decoded per step (for mitigating accumulation errors) leads to redundant sampling passes.
As a result, Gumbel-max has evolved to a heavily tuned time variant trick with unstable performance, particularly when scaled to large-vocabulary codebooks.
The above factors, rather than quantization alone, induce the performance gap between discrete and continuous models.

In this study, we propose a discrete diffusion model with an elaborate rehashing noise design, Fig.~\ref{fig1}(lower).
Our approach, termed \textbf{ReDDiT}, addresses the limitations of the uni-mask design by redefining absorbing states towards larger representational capacity, through enriching the potential paths that latent variables can traverse during diffusion.
Specifically, we expand the masks to multiple indices along with the codebook and randomize them during data corruption.
A rehash sampler is also derived with principled discrete diffusion theories to reverse the diffusion path for generation, guaranteeing high diversity and low discrepancy of the sampling process.
We demonstrate that this rehashed noise facilitates learning a superior and regularized expressiveness, while eliminating reliance to hyper-parameterized randomness during sampling.

We further revisit the commonly used discrete diffusion objective and update it with empirical modifications.
By adopting an improved ELBO~\cite{sahooSimpleEffectiveMasked2024, shiSimplifiedGeneralizedMasked2025} with representation alignment (RepA)~\cite{RePA24} loss, we optimize the training efficiency and substantially improve the generation quality of discrete generative models.
Moreover, ReDDiT aligns with recent advances in discrete flow matching~\cite{gatDiscreteFlowMatching2024, shaulFlowMatchingGeneral2024}, enabling token refreshment during sampling without training post-correction models~\cite{lezamaImprovedMaskedcritic2022}.

\section{Methodology}
\label{sec:method}
For self-containment, we first review the DDM theory in Sec.~\ref{sec:prelim}. We then reformulate its diffusion dynamics and introduce rehashing noise for ReDDiT in Sec.~\ref{sec:reform}. We finally discuss connection and comparison with other discrete diffusion models in Sec.~\ref{sec:discuss}.

\subsection{Preliminary: Discrete Diffusion Model}
\label{sec:prelim}
DDM defines a forward process over discrete variables by gradually corrupting the image tokens to absorbing states (masks) through a continuous-time Markov process.
Assume that the data consists of tokens from a finite vocabulary $\mathcal{V}$. $x \in \mathcal{V}^L$ is a sequence of tokens ($e.g.$, an image tokenized into indices) with length $L$. We denote the clean data as $x_{t=0}$ ($x_0$ for short), and noise it gradually as $t\rightarrow 1$.
DDM defines an absorbing token $\mathbf{m} \in \mathcal{V}$, such that once a token is noised to $\mathbf{m}$ it remains unchanged.
At the terminal time $t=1$, \( x_t \) fully transits to $\mathbf{m}^L$, which means $x_1^{i=1\sim L}=\mathbf{m}$.

Let \( \alpha_t \) be the noise scheduler (a monotonically decreasing survival function that satisfies \(\alpha_0=1, \alpha_1=0\) ).
For \( 0 \leq s < t \leq 1 \), the forward corruption process is governed by a continuous-time transition kernel \( q(x_t^i|x_s^i) \) at the $i$-th element, as
\begin{equation}\label{eq1}
q(x_t^i|x_s^i) = 
\begin{cases}
1 - \alpha_{t|s}, & \text{if } x_t^i = \mathbf{m}, x_s^i\ne\mathbf{m} \\
\alpha_{t|s}, & \text{if } x_t^i = x_s^i, x_s^i\ne\mathbf{m} \\
1, & \text{if } x_t^i = x_s^i, x_s^i=\mathbf{m} \\
0, & \text{otherwise}
\end{cases}
,\quad \alpha_{t|s} =\frac{\alpha_t}{\alpha_s}.
\end{equation}
%
Denoting $q$ as the transition kernel and $\text{Cat}(\cdot;\pi)$ the categorical distribution determined by probability $\pi$, the corrupted data distribution at time $t$ is written as 
\begin{equation}
    x_t \sim q(x_t|x_0), q(x_t|x_0) = \text{Cat}(x_t;\alpha_t x_0 + (1-\alpha_t)\mathbf{m}^L).
\end{equation}
The generative model learns the reverse process \( p_\theta(x_s|x_t) \), which denoises sample \( x_t \) at arbitrary time \( t \in (0, 1] \) to a less noised state \( x_s \) at time \( s < t \).
Denoting $\delta(x_t^i,\textbf{m})$ as the indicator function that only computes on masked tokens, and $\alpha_t'=\frac{\mathrm{d}\alpha_t}{\mathrm{d}t}$, the learning objective is derived ~\cite{shiSimplifiedGeneralizedMasked2025} as
\begin{equation}\label{ddmloss-1}
    \mathcal{L}_{\text{DDM}} = -\mathbb{E}_{x_0,\ x_t} \int_{t=0}^{t=1}[\frac{\alpha_t'}{1-\alpha_t}\sum_{i=1}^L\delta(x_t^i,\textbf{m})\log p_\theta(x_0^i|x_t)]\mathrm{d}t \ .
\end{equation}
For a linear scheduler, Eq.~\ref{ddmloss-1} is simplified via variable substitution~\cite{sahooSimpleEffectiveMasked2024} to an equivalent form, as
\begin{equation}\label{ddmloss}
    \mathcal{L}_{\text{DDM-linear}} = -\mathbb{E}_{t,\ x_0,\ x_t} [\frac{1}{t}\sum_{i=1}^L\delta(x_t^i,\textbf{m})\log p_\theta(x_0^i|x_t)]\ .
\end{equation}
For conditional generation, class information \( c \) (\textit{e.g.}, labels or text prompts) is introduced to the denoising model as additional input.
Following classifier-free guidance~\cite{hoClassifierFreeDiffusionGuidance2022}, the model is trained with a random drop of labels, and the prediction is interpolated at inference, as
\begin{equation}
    \hat{p}_\theta(x_t, c) = \ p_\theta(x_t, \varnothing) + w \cdot (\, p_\theta(x_t, c)-\ p_\theta(x_t, \varnothing)),
\end{equation}
where $\varnothing$ is the dropped label and \( w \geq 0 \) controls the guidance strength.

%
\subsection{Discrete Diffusion with Rehashing Noise}
\label{sec:reform}
\begin{figure}[t]
\centering
\includegraphics[width=\linewidth]{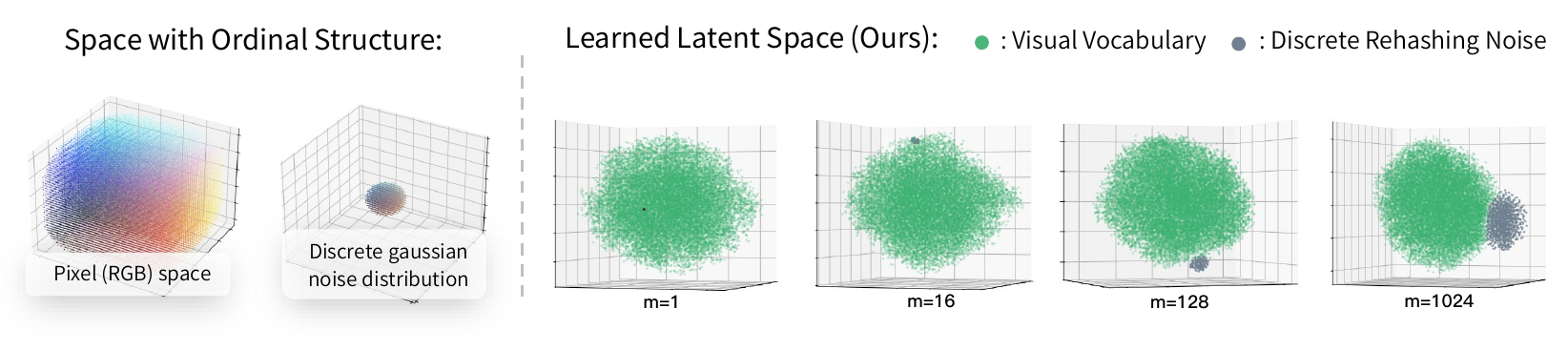}
\caption{\textbf{Visualization of pixel and latent spaces.} $m$ denotes the number of enriched noise indices. 3D t-SNE plot (right) is used solely to separate valid and noise tokens for clustering illustration.}
\label{fig:noise-vis}
\end{figure}

The ordinal structure inherent in discrete data provides a valuable inductive bias for designing transition kernels in diffusion dynamics. Prior studies~\cite{austinStructuredDenoisingDiffusion2023,campbellContinuousTimeFramework2022} show that assigning higher transition probabilities to neighboring pixel values—forming a \textit{discrete Gaussian-like noise}—outperforms the single absorbing state approach on pixel-level datasets like CIFAR-10. 
However, when using visual tokenizers, the structure of discretized latents is learned rather than pre-defined, making such ordinal assumptions inapplicable.
This insight motivates us to extend conventional mask tokens to a set of indices, and reverse the diffusion path with noise rehashing.
This design allows the model to optimize its embedding space during training, enhancing its ability to model flexible and data-driven noise structures. We visualize the learned distributions in Fig.~\ref{fig:noise-vis} (right).
\paragraph{Reformulation.}
Given $d$ categories, let $\mathbf{e}_i\in \mathbb{R}^{d}$ be its one-hot vector where the $i$-th value is $1$.
We denote $ \mathcal{E}=\{ \mathbf{e}_i \in \mathbb{R}^{d} \mid i = 1, \ldots, d \}$ as the basis of a categorical distribution (known as $d$-simplex), 
and a basis for absorbing states with capacity $m$:  $\mathcal{M}=\{ \mathbf{m}_j \in \mathbb{R}^{m}\mid j = 1, \ldots, m \}$. 
The sum of $\mathcal{E}$ and $\mathcal{M}$ can be denoted as 
\begin{equation}
\mathcal{V}_{(d,m)} \triangleq \left\{ \mathbf{v}_{(i,j)} \in \mathbb{R}^{d+m} \,\middle|\, 
\mathbf{v}_{(i,j)} = 
\begin{cases}
\mathbf{e}_i \oplus \mathbf{0}_m, & \text{for } i = 1, \ldots, d,\ j=0 \\
\mathbf{0}_d \oplus \mathbf{m}_j, & \text{for } j = 1, \ldots, m, \ i=0
\end{cases}
\right\}.
\end{equation}
We further denote the subspace $\mathcal{E}_d, \ \mathcal{M}_m\in \mathcal{V}_{(d,m)}$ which contain \textit{either} valid or mask tokens, as
\begin{equation}
\mathcal{E}_d = \left\{ \mathbf{v}_{(i,0)} \in \mathcal{V}_{(d,m)} \,\middle|\, i = 1, \ldots, d \right\}, \ \mathcal{M}_m = \left\{ \mathbf{v}_{(0,j)} \in \mathcal{V}_{(d,m)} \,\middle|\, j = 1, \ldots, m \right\}.
\end{equation}
To exploit visits across all the possible paths, we rewrite the transition kernel defined by Eq.~\ref{eq1} as
\begin{equation}\label{rewrite}
    q(x_t^i|x_s^i) = 
    \begin{cases}
    1 - \alpha_{t|s}, & \text{if } x_t^i \in \mathcal{M}_m,\ x_s^i\notin\mathcal{M}_m \\
    \alpha_{t|s}, & \text{if } x_t^i = x_s^i,\ x_s^i\notin\mathcal{M}_m \\
    1/m, & \text{if } x_t^i \in\mathcal{M}_m, \ x_s^i\in\mathcal{M}_m \\
    0, & \text{otherwise}.
    \end{cases}
\end{equation}
With above definitions, we reformulate the diffusion process of $x$ as a \textbf{transition from $\mathcal{E}_d$ to $\mathcal{M}_m$}. We train the model by feeding it with corrupted data, of which the distribution is inferred as $x_t \sim \text{Cat}(x_t;\alpha_{t}x_0 + (1-\alpha_{t})\text{U}(\mathcal{M}_m^L))$, where $\text{U}(\mathcal{M}_m^L)$ is the uniform distribution upon $\mathcal{M}_m^L$. 
\paragraph{Rehash Sampling.}
To generate a sequence of length $L$, the reverse process starts with $x_1 \sim \text{U}(\mathcal{M}_m^L)$.
The subsequent latents $x_t$ are generated by discretizing the reverse timeline $T$ to $K$ steps. 
We denote this schedule as $T^{1:K+1}$ such that $T^1=1$ and $T^{K+1} =\varepsilon$, with $\varepsilon$ being an arbitrarily small positive constant.
The reverse process is deduced from the formulation, as
\begin{equation}\label{reverse}
    q_{s|t}^i = q(x_{s}^i|x_{t}) = 
    \begin{cases}
    1, & \text{if } x_s^i = x_t^i, \ x_t^i \notin \mathcal{M}_m\\
    \frac{1-\alpha_s}{m(1-\alpha_t)}, & \text{if } x_s^i \in \mathcal{M}_m, \ x_t^i \in \mathcal{M}_m\\
    \frac{\alpha_s-\alpha_t}{1-\alpha_t}p_\theta^i(x_t), & \text{if } x_s^i \notin \mathcal{M}_m, \ x_t^i \in \mathcal{M}_m  \\
    0, & \text{otherwise.} \\
    \end{cases}
\end{equation}
Comparing with MVTM sampler in Alg.~\ref{algo-mvtm}, our rehash sampler is shown in Alg.~\ref{algo-rehash}. Similar to MDLM~\cite{sahooSimpleEffectiveMasked2024}, we apply \texttt{torch.multinomial} (Multnm. in step 10) for low-discrepancy\footnote{Instead of dividing $\alpha_{s|t}$ and assigning these probabilities to each mask vocabulary. We merge the probabilities at step 9 to 
keep an overall noise sampling probability, as small values might be truncated.} categorical sampling. 

\begin{minipage}{0.48\textwidth}
\begin{algorithm}[H]
\caption{MVTM Sampling}
\label{algo-mvtm}
\begin{algorithmic}[1]
\State \textbf{Inputs:} label $c$, scheduler $\alpha_t$, length $L$, 
\State \textbf{Settings:} number of steps $K$, $G(t)$, $\mathcal{G}$
\State Initialize: $x_1 \gets \mathcal{M}_1^L$, $t \gets 1$.
\For{$k = 1$ to $K$}
    \State $t \gets \frac{K - k+1}{K}, s \gets \frac{K - k}{K}$
    \State $p_{\text{score}}\gets f_\theta(x_t, c) + G(t)\cdot \mathcal{G}$
    \State $x_{\text{pred}} \gets \text{argmax}(p_{\text{score}})$
    \Comment{Predict-all}
    \State $x_s \gets \text{where}(x_t = [\text{m}], x_{\text{pred}}, x_t)$
    \State $p_{\text{conf}} \gets p_{\text{score}} + G(t)\cdot \mathcal{G}$
    \State $m_\text{re} \gets \text{argsort}(p_{\text{conf}})[1 : L \cdot (1 - \alpha_s)]$
    \State $x_s \gets \text{where}(m_\text{re}, [\text{m}], x_s)$
    \Comment{Re-mask}
\EndFor
\State \textbf{Return:} fully unmasked sequence $x_0$
\end{algorithmic}
\end{algorithm}
\end{minipage}
\hfill
\begin{minipage}{0.48\textwidth}
\begin{algorithm}[H]
\caption{Rehash Sampling}
\label{algo-rehash}
\begin{algorithmic}[1]
\State \textbf{Inputs:} label $c$, scheduler $\alpha_t$, length $L$.
\State \textbf{Settings:} number of steps $K$.
\State Initialize: $x_1 \sim \text{U}(\mathcal{M}_m^L)$, $t \gets 1$, $T^{1:K}$.
\For{$k = 1$ to $K$}
    \State $t \gets T^k, s \gets T^{k+1}$
    \State $x_t \gets \text{where}(x_t \in\mathcal{M}_m,\text{U}(\mathcal{M}_m^L), x_t)$
    \State $p \gets \pi_\theta(x_t, c)$
    \State $q_{s|t}\gets \frac{\alpha_s - \alpha_t}{1- \alpha_t} \cdot p +\delta\cdot\frac{1-\alpha_s}{1-\alpha_t}$
    \State $x_{\text{pred}} \gets \text{Multnm.}(q_{s|t})$
    \Comment{w/ masks}
    \State $x_s \gets \text{where}(x_t \in \mathcal{M}_m, x_{\text{pred}}, x_t)$
\EndFor
\State \textbf{Return:} fully unmasked sequence $x_0$
\end{algorithmic}
\end{algorithm}
\end{minipage}

The random nature of absorbing states inspires a rehash operation: we shuffle these tokens at the beginning of each step by $x_t \gets \text{where}(x_t \in\mathcal{M}_m,\text{U}(\mathcal{M}_m^L), x_t)$. Proof to Eq.\ref{reverse} is included in Appendix.~\ref{proof-appen}.
%
\subsection{Discussion}
\label{sec:discuss}
\begin{figure}[h]
\centering
\includegraphics[width=\linewidth]{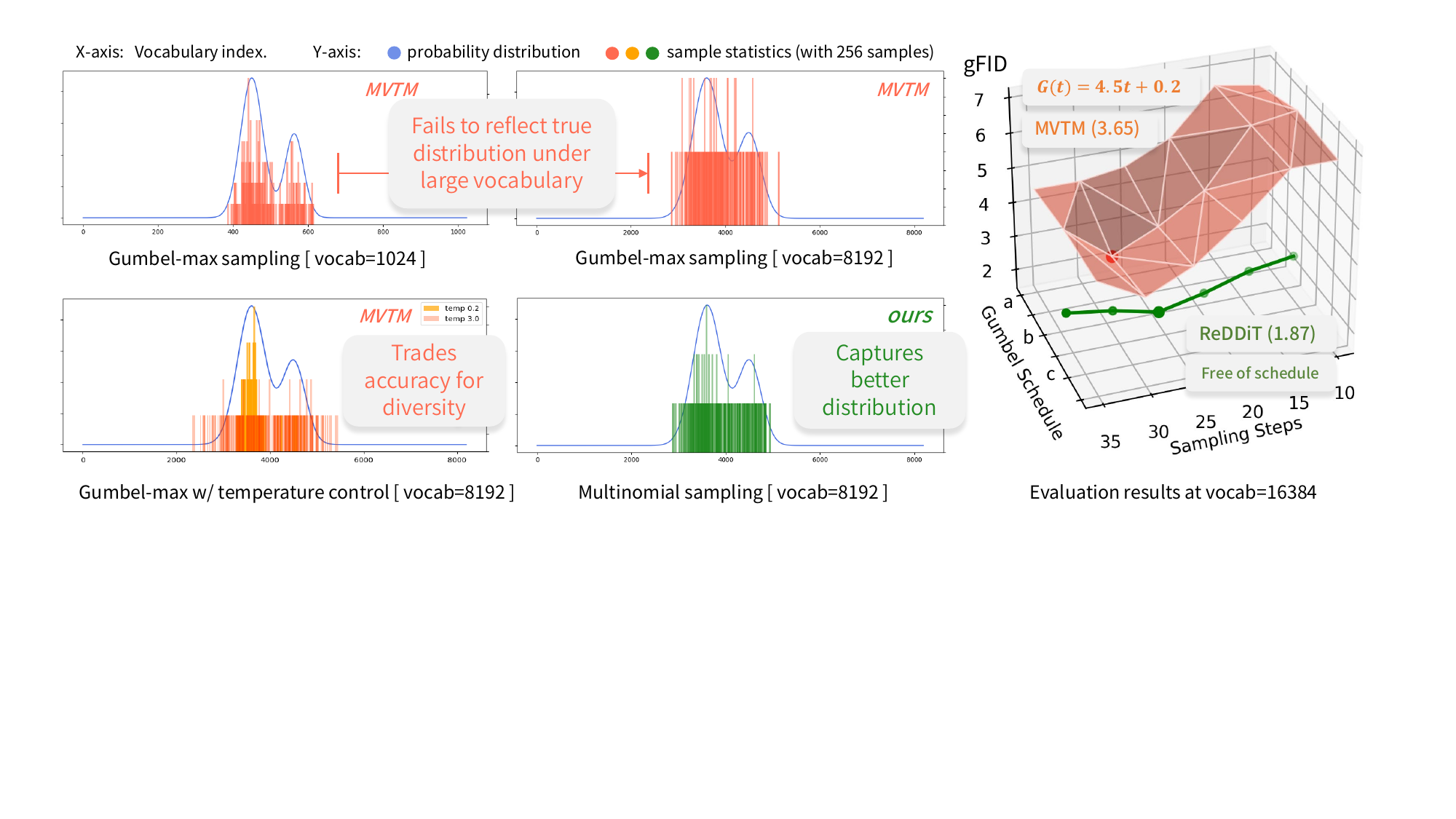}
\caption{\textbf{Sampler comparison.} \textit{Left}: Gumbel-max is theoretically equivalent to our method, yet it struggles to reflect the true distribution under limited sample passes.
The multinomial approach captures the distribution more accurately.
\textit{Right}: our model achieves lower gFID across different sampling steps without tuning Gumbel-max, indicating more efficient and faithful sampling. \textit{a, b, c} refer to three uniformly sampled $G(t)$ set for MVTM sampling. See supplementary for experimental codes. (This figure is best viewed in color)}
\label{fig:gumbels}
\end{figure}
\paragraph{Comparison with MVTM.}
Masked visual token models (MVTMs) borrow the objective 
\begin{equation}\label{mvtmloss}
    \mathcal{L}_{\text{MVTM}} = -\mathbb{E}_{t,\ x_0,\ x_t} \sum_{i=1}^L\delta(x_t^i,\textbf{m})\log p_\theta(x_0^i \mid x_t),
\end{equation}
from masked language models~\cite{devlinBERTPretrainingDeep2019} and predict on masked tokens with a maximum likelihood.
Besides the reformulated corruption (Eq.~\ref{rewrite}) and reverse process (Eq.~\ref{reverse}), ReDDiT differs in the following aspects: (\textit{i}) the training objective (Eq.~\ref{ddmloss}), which is derived from DDM, providing better theoretical and empirical results. 
(\textit{ii}) it can easily sample with a arbitrarily discretized timeline, while MVTM couples training and inference, restricting its sampling flexibility; 
(\textit{iii}) the rehash sampler (Alg.~\ref{algo-rehash}) includes absorbing states in categorical sampling with lower discrepancy, different from MVTM's predict-remask sampler with time variant intensity $G(t)$ over Gumbel noise $\mathcal{G}$ (Alg.~\ref{algo-mvtm})
\footnote{The logits corresponding to previously restored tokens' indices are manually set to infinity for both methods, so that they will not be noised again in the following steps. 
This leads to an implementation of any-order auto-regressive model~\cite{ouYourAbsorbingDiscrete2025} if DDM's decoded tokens per step is limited to $1$.}.
Gumbel-max suffers from numerical inaccuracy~\cite{zhengMaskedDiffusionModels2025} and we noitice that it becomes worse on large vocabulary (Fig.~\ref{fig1},~\ref{fig:gumbels} with our reproduced results), which limits MVTM's potential.

\paragraph{Relationship to DFM.}
%
Discrete flow matching (DFM)~\cite{gatDiscreteFlowMatching2024} introduces a transition process based on masked tokens.
Its training objective was initially designed as the masked token loss (~\ref{mvtmloss}), and evolved to a time-weighted cross-entropy loss~\cite{shaulFlowMatchingGeneral2024} for generalized diffusion paths, which is similar to ours.
The similarity enables a direct comparison between the DFM sampler and our rehash sampler using the same trained model weights.
We notice that it generally requires more steps to reach optimal results, as the DFM sampler offers a refinement mechanism via token-wise updates.
Since the gradual decoding method is shared, we can integrate certain DFM steps into our sampling procedure for refinement.
This leads to $\sim$ 0.1 gFID improvement on ImageNet-1K.
{Refer to Appendix~\ref{appen-dfm} for details.}
%
%
\section{Experiment}
\label{sec:exp}

\subsection{Implementation}
\label{sec:detail}

\paragraph{Datasets.}
The experiments are conducted on ImageNet-1K~\cite{imageNet2009}, which consists of 1000 categories, 1281167 images and are cropped to resolution $256 \times 256$ for training.
The generation quality is evaluated using Fréchet Inception Distance (FID)~\cite{heusel2017gans} and the Inception Score (IS)~\cite{IS2016}.
FID measures the distance between the distributions of generated and real images in the feature space of a pre-trained Inception network,
while IS evaluates both the confidence and diversity of generated images by analyzing predicted label distribution. We compute generation FID (gFID$\downarrow$)\footnote{The gFID is used as the quality metric for generative models' performance, while rFID refers to the reconstruction quality of a visual tokenizer.} and IS$\uparrow$ on 50k generated samples. 
\paragraph{Pre-processing.}
Following the setting in LlamaGen~\cite{sunllamagen2024}, we apply the ten-crop augmentation on images, and use pre-trained tokenizers to convert them to discrete tokens.
We pick IBQ-f16~\cite{shiScalableImageIBQ2025} tokenizer as default for its scalable and promising performance in generation tasks, which uses a $16\times16$ downsampling ratio and converts a $256\times256$ image into 256 discrete tokens. The tokenizer has a codebook with 16384 entries.
The LlamaGen-f16 (used in Tab.~\ref{table:vqinfo}) and LlamaGen-f8 tokenizer~\cite{sunllamagen2024} (used in Tab.~\ref{tab_perfor_com}) are also used for comparison with recent discrete generation methods.  All tokenizers are used out-of-the-box without modification.

\paragraph{Representation Alignment.}
Recent study~\cite{RePA24} has shown that the alignment of intermediate representations between diffusion transformers and vision encoders accelerates training convergence of diffusion models.
Accordingly, the alignment is designed as a regularization term with $\lambda=0.5$.
We extract diffusion transformer's \textit{8}-th layer intermediate feature $\textbf{h}^{[n]}(x_t)$ and align it with the original image's dinov2-b~\cite{oquabDINOv2LearningRobust2024} encoded features $f_{\text{enc}}(x_0^{\text{ori}})$.
The intermediate features are projected by a small trainable MLP $h_{\varphi}$. The $\text{sim}(\cdot,\cdot)$ computes the mean of element-wise cosine similarity between embeddings, as
\begin{equation}\label{totalloss}
    \mathcal{L}_{\text{total}} = \mathcal{L}_{\text{DDM-linear}} + \lambda \mathcal{L}_{\text{RepA}}, \ \ \mathcal{L}_{\text{RepA}} = -\mathbb{E}_{x, \ t}[\ \text{sim}(f_{\text{enc}}(x_0^{\text{ori}}),\ h_{\varphi}(\textbf{h}^{[n]}(x_t)))\ ]\ .
\end{equation}
This alignment was proposed for continuous diffusion models, and we firstly validate that it's also suitable for training discrete models. However, from our observation, as a training acceleration technique, RepA \textbf{does not} provide relative performance gain if training sufficiently (like for 1M steps as most diffusion models do) for discrete latents. We only use RepA to improve training efficiency and probe the inner dynamics through training as in Fig.~\ref{fig:noise-cap}.

\paragraph{Training and Evaluation.}
The proposed model is based on DiT~\cite{peeblesScalableDiffusionModels2023} architecture, with reference to its discrete prediction version~\cite{sahooSimpleEffectiveMasked2024}.
2D-RoPE~\cite{suRopeFormerEnhancedTransformer2023} and min-SNR~\cite{hangsnrEfficientDiffusion2023} are applied for training efficiency.
The model is optimized using the AdamW optimizer with a cosine decay.
Training is conducted for 500k iterations on 8 NVIDIA H100 GPUs with a global batch size 1024. Class-conditional training is enabled using class embeddings and a drop-rate of 0.1 for generation with classifier-free guidance. Details are provided in Appendix~\ref{appen-details}. 
\subsection{Performance and Comparison}
\label{sec:compare}
We compare the proposed ReDDiT model with other generative models on the ImageNet-1K $256\times256$ in Tab.~\ref{tab_perfor_com}.
The IBQ tokenizer is used for the default L and XL models. We also utilize LlamaGen-f8 with 128 noise capacity to evaluate its high-resolution potentials (noted as ReDDiT-XL$\rm _{f8}$). We use a linear increasing guidance following the common practice of ~\cite{gaoMDTv2MaskedDiffusion2024}.
\begin{table}[t]
\centering
\caption{\textbf{Performance comparison on class-conditional ImageNet 256$\times$256.} 
%
Look-up free quantizers are beyond the scope of this paper. \textit{ft.}(in gray) indicates that the decoder is fine-tuned for quantized latents. Wall-clock inference time relative to ReDDiT-XL is reported.}
\scalebox{0.8}{
\begin{tabular}{l|lcccccccc}
\toprule
\multirow{2}{*}{Type} & \multirow{2}{*}{Model} & \multicolumn{2}{c}{Tokenizer} & \multicolumn{5}{c}{Generator} \\
\cmidrule(lr){3-4}\cmidrule(lr){5-9}
&  & \#tokens & codebook & gFID$\downarrow$ & IS$\uparrow$ & \#Params & \#Steps &Time\\
\midrule
\multirow{4}{*}{Diff.} & LDM-4~\cite{LDM4} & 4096×3 & - & 3.60 & 247.7 & 400M & 250 & -- \\
& DiT-XL/2~\cite{peeblesScalableDiffusionModels2023} & 1024×4 & - & 2.27 & 278.2 & 675M & 250 & 18 \\
& SiT-XL~\cite{maSiTExploringFlow2024a}& 1024×4 & - & 2.42 & 238.5 & 675M & 30 & 2 \\
& SiT-XL w/ Solver~\cite{wangDifferentiableSolverSearch2025} & 1024×4 & - & 2.24 & 244.1 & 730M & 15 & 1.2 \\
\midrule
\multirow{6}{*}{AR} & Taming-VQGAN~\cite{TamingVQGAN} & 256 & 1024 & 15.78 & 74.3 & 1.4B & 256 & 8\\
& RQ-Transformer~\cite{RQTransformer23} & 256 & 16384 & 7.55 & 134.0 & 3.8B & 64 & 8.5\\
& ViT-VQGAN~\cite{ViTVQGAN22} & 1024 & 8192 & 4.17 & 175.1 & 1.7B & 1024 & >10\\
& LlamaGen-3B~\cite{sunllamagen2024} & 576 & 16384 & 2.18 & 263.3 & 3.1B & 576 & 20\\
& RandAR-XXL~\cite{pangRandARDecoderonlyAutoregressive2024} & 512 & 16384 & 2.15 & 322.0 & 1.4B & 88 & 4\\
& VAR-\textit{d}30~\cite{tian_VAR_2024} & 680 & 4096 & 1.97 & 334.7 & 2.0B & 10 & 0.5\\
\midrule
\multirow{2}{*}{MVTM} & MaskGIT~\cite{changMaskGITMaskedGenerative2022} & 256 & 1024 & 6.18 & 182.1 & 227M & 8 & 0.2\\
& MaskGIL-XXL~\cite{xinResurrectMaskAutoRegressive2025} & 256& 16384 & 3.71 & 303.4 & 1.4B & 8 & 0.8 \\
& \color{gray} TiTok-S-128$_{ft.}$~\cite{yutitok2024} & \color{gray} 128 & \color{gray} 4096 & \color{gray} 1.97 & \color{gray} 281.8 & \color{gray} 287M & \color{gray} 64 & \color{gray} 1.6\\
\midrule
\multirow{4}{*}{DDM} & ITM~\cite{huMASKAllYou2024}  & 1024 & 16384 & 5.30 & 183.0 & 546M & 100 & 3\\
& ReDDiT-L (ours) & 256& 16384 & 2.13 & 294.7 & 346M & 20 & 0.5 \\
& ReDDiT-XL (ours)  & 256 & 16384 & 1.74 & 313.6 & 675M & 32 & 1\\
& ReDDiT-XL$\rm _{f8}$ (ours)  & 1024 & 16384 & \bf 1.61 & 318.5 & 675M & 64 & 2\\
\bottomrule
\end{tabular}
}
\label{tab_perfor_com}
\end{table}

\begin{table}[t]
\caption{\textbf{Comparison of models with the same tokenizer.}   Reconstruction FID (rFID) indicates the tokenizer's reconstruction quality from its quantized codes. Dim denotes codebook dimension. AR model's gFID are indexed from their original report.}
\label{table:vqinfo}
\centering
\scalebox{0.9}{
\begin{tabular}{lcccccc}
\toprule
\multirow{2}{*}{Model} & \multicolumn{3}{c}{VQ Tokenizer Info.} & \multicolumn{2}{c}{Generator} \\
\cmidrule(lr){2-4} \cmidrule(lr){5-6}
 & Identity & rFID & dim & \#Params & gFID$\downarrow$ \\
\midrule
LlamaGen-L$\rm _{AR}$~\cite{sunllamagen2024} &\multirow{3}{*}{LlamaGen-f16~\cite{sunllamagen2024}} & \multirow{3}{*}{2.19} & \multirow{3}{*}{8} & 343M & 3.80   \\
RandAR-L$\rm _{AR}$~\cite{pangRandARDecoderonlyAutoregressive2024} &  &  & & 343M & 2.55  \\
Ours$\rm _{DDM(ReDDiT-L)}$ &  &  & & 346M & 2.41  \\
\midrule
IBQ-B$\rm _{AR}$~\cite{shiScalableImageIBQ2025} & \multirow{2}{*}{IBQ-tokenizer~\cite{shiScalableImageIBQ2025}} & \multirow{2}{*}{1.37} & \multirow{2}{*}{256} & 343M & 2.88   \\
Ours$\rm _{DDM(ReDDiT-L)}$  &  &  & & 346M & 2.13  \\
\bottomrule
\end{tabular}
}
\end{table}

\paragraph{Generation Quality.}
As shown in Tab.~\ref{tab_perfor_com}, ReDDiT achieves the best performance among the compared discrete models.
It outperforms the baseline (MaskGIT~\cite{changMaskGITMaskedGenerative2022}) with significant margins (gFID: 2.13 vs 6.18 and IS: 294.7 vs. 182.1).
It also outperforms the recent DDM method~\cite{huMASKAllYou2024} and TiTok-S-128~\cite{yutitok2024}, which is extensively fine-tuned on quantized latents.
Compared with continuous diffusion models, ReDDiT exhibits on-par efficiency and performance, showing great potential for discrete generation.
Note that the performance is achieved with a codebook size of 16384, validating ReDDiT's effectiveness for large-vocabulary codebooks.
\paragraph{Efficiency.}
ReDDiT is born with the high-efficiency advantage of discrete diffusion models, comparing with AR models.
As shown in Tab.~\ref{tab_perfor_com}, the inference time of ReDDiT is slightly longer than MaskGIT, while the performance is overwhelming.
Without acceleration techniques, ReDDiT achieves a competitive performance which AR and traditional diffusion models use more than 250 steps to achieve.
Notably, when armed with recent efforts that tailored KV-Cache~\cite{liuDLLMCacheAcceleratingDiffusion2025} for discrete diffusion models, ReDDiT's inference can be further boosted (not included in the main paper for fair comparison). See Appendix~\ref{acc} for details.

Besides the major comparison, we also conduct an experiment that utilizes the identical tokenizer in previous AR models and validate our method's effectiveness. As can be seen in Tab.~\ref{table:vqinfo}, ReDDiT outperforms AR methods in generation tasks across different tokenizers. Note that this comparison is to demonstrate diffusion model's potential on discretized latents, and current representation alignment methods are inapplicable to AR models due to their unidirectional attention design.
\subsection{Determining Noise Capacity}
\begin{figure}[h]
\centering
\includegraphics[width=\linewidth]{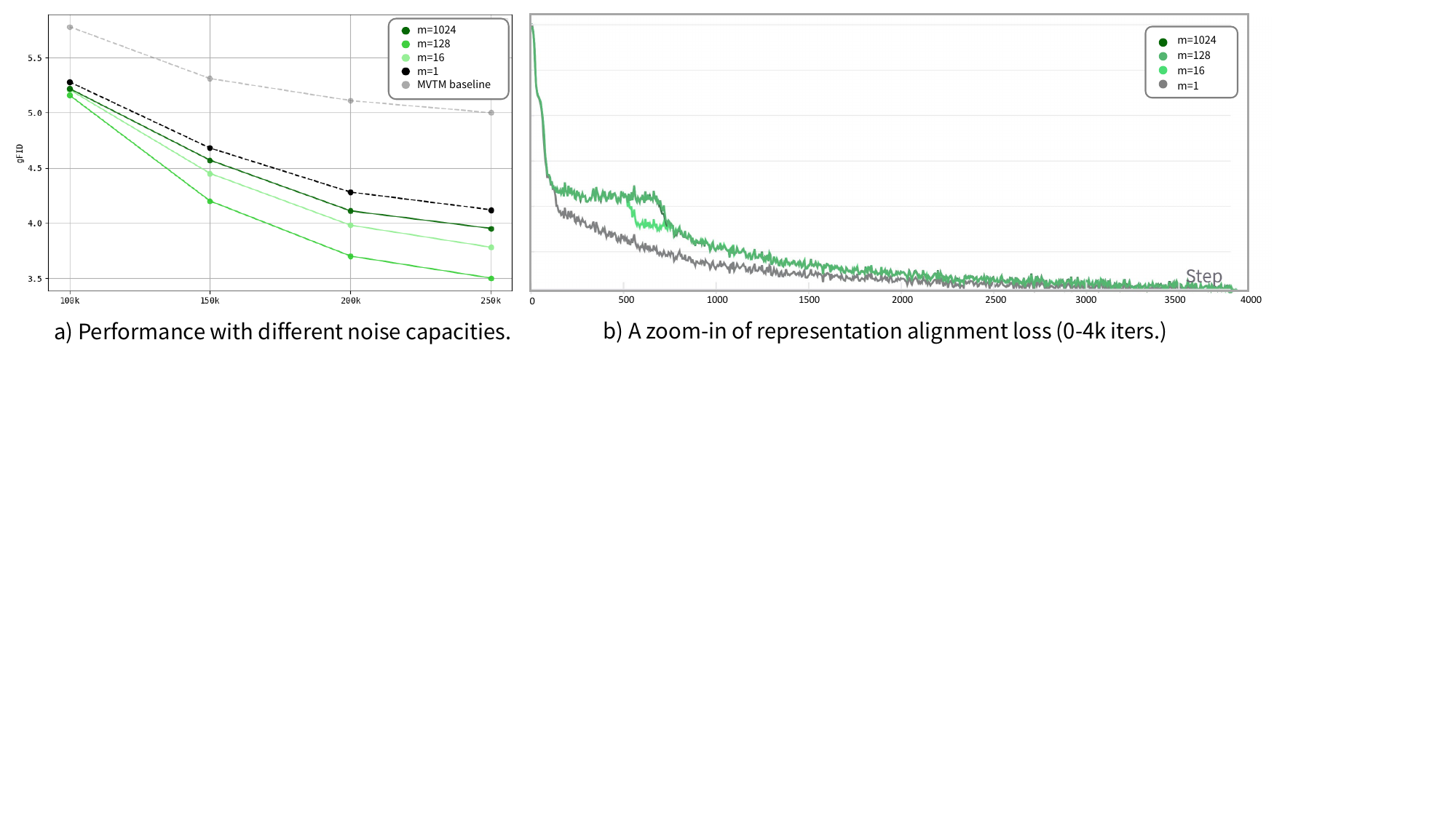}
\caption{\textbf{Comparison of noise capacities.} We re-implemented training with MVTM and ReDDiT design with the same training recipe (LlamaGen-f16 as visual tokenizer, with codebook size $16384$). The generation quality and representation alignment trends are visualized. }
\label{fig:noise-cap}
\end{figure}
The reformulated discrete diffusion dynamics defines transitioning from $\mathcal{E}_d$ to $\mathcal{M}_m$.
Under this setting, it is necessary to empirically determine the optimal value of $m$ for a fixed tokenizer with vocabulary size $d$, as the latent representations learned by VAEs are variant.
We keep the training setup fixed and conduct experiments \textit{w.r.t.} the noise capacity $m$.
We also visualize $\mathcal{L}_{\text{RepA}}$, which captures the degree of representation alignment~\cite{RePA24} within the transformer.

The alignment loss visualization shows that increasing the number of absorbing states introduces greater randomness, initially making predictions more difficult due to confusion with valid tokens. However, this gap narrows as training progresses, and the model converges to a similar alignment lower bound, suggesting effective representation learning across different configurations.

As shown in Fig.~\ref{fig:noise-cap} (left), generation quality improves with increasing noise capacity initially. The LlamaGen-f16 tokenizer achieves peak performance at $m=128$, while the IBQ tokenizer performs best at $m=1024$.
We attribute this to the codebook design: the lower dimensional LlamaGen-f16 codebook produces more compact latents, which also determines its smaller noise endurance.
\subsection{Ablation Study}
\label{sec:abl}
%
%
Unless specified, all the models are trained on ImageNet $256 \times 256$ under the default settings for 100k iterations for fair comparison. We use a constant guidance scale of 2.0 and 20 steps for generation, and report gFID $\downarrow$ computed on 50K samples. Precision (Prec.$\uparrow$) and Recall (Rec.$\uparrow$) are also reported in general design for direct diversity comparison.
\paragraph{Sampling Timeline.}
\label{sec:timewarp}
\begin{wrapfigure}{r}{6cm}
\centering
\includegraphics[width=0.43\textwidth]{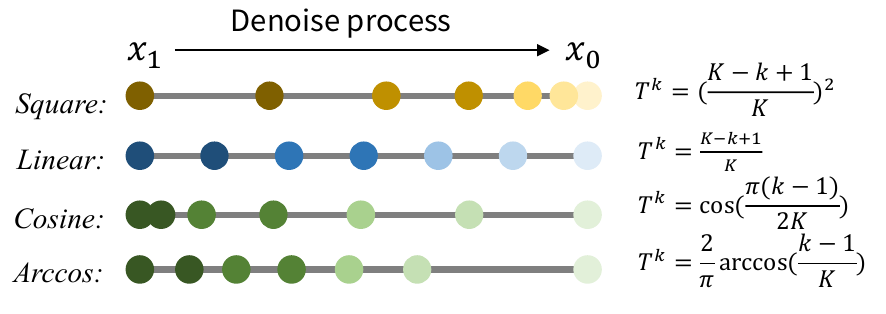}
\caption{\footnotesize Illustration of discretized timeline with $K=7$. The slow-to-fast sampling works better than linear schedules.}
\label{fig:timeline}
\end{wrapfigure}

Recovering complete information from noise remains critical to diffusion-based models~\cite{luDPMSolverFastSolver2023,wuRDPMSolveDiffusion2024}.
Recent work shows MVTM's non-linear scheduler for training is less critical when using high-capacity tokenizers.
Evidence of time-invariance in DDMs~\cite{sahooSimpleEffectiveMasked2024,shiSimplifiedGeneralizedMasked2025} further supports decoupling training from sampling. In our experiments, a linear scheduler with constant signal-to-noise ratio decay, yields optimal training dynamics. Among the timeline discretization tested, Fig.~\ref{fig:timeline}, the \textit{cosine} schedule is employed for our ReDDiT model for best performance in Tab.~\ref{subtab:sample}.
\paragraph{General Design.}
\begin{table}[ht]
\centering
\caption{\textbf{Ablated Design choices.} ReDDiT-L is trained for 100k iters. Final setting denoted in \textcolor{gray}{gray}.}
\begin{minipage}[t]{0.7\textwidth}
\centering
\subcaption*{\footnotesize (a) General model design}\label{subtab:general}
\scalebox{0.9}{
\label{tab:ablation}
\begin{tabular}{lcccc}
\toprule
Train Config & Sample Config & gFID & Prec. & Rec.  \\
\midrule
MVTM + RepA loss  & MVTM sampler & \textit{6.83} & 0.75 & 0.39\\
Switch to objective (\ref{totalloss}) & MVTM sampler & 6.23 & 0.77 & 0.41\\
\ \ \ same as above & Rehash sampler & \textit{5.75} & 0.78 & 0.45\\
\ + 2D-RoPE + min-SNR & Rehash sampler & 5.51 &0.79 & 0.45\\
\rowcolor{gray!20} \ \ \  same as above & + DFM refine & 5.40& 0.81 & 0.52\\
\bottomrule
\end{tabular}
}
\end{minipage}
\hfill
\begin{minipage}[t]{0.28\textwidth}
\centering
\subcaption*{\footnotesize (b) Sampling timeline}\label{subtab:sample}
\scalebox{0.9}{
\begin{tabular}{ccc}
\toprule
Steps & Timeline & gFID \\
\midrule
20 & \textit{linear} & 7.18 \\
32 & \textit{linear} & 6.43 \\
20 & \textit{arccos} & 5.04 \\
20 & \textit{square} & 7.39 \\
\rowcolor{gray!20} 20 & \textit{cosine} & 4.91 \\
\bottomrule
\end{tabular}
}
\end{minipage}
\end{table}
We ablate the general choices of ReDDiT, which starts with a re-trained MVTM baseline methods (with LlamaGen-f16 and RepA for faster convergence as default) in Tab.~\ref{subtab:general}. The applied techniques like 2D-RoPE are also ablated with re-training. As shown, through the revised objective and our proposed sampler, ReDDiT alone improves FID by $\sim1.0$ compared to the baseline model. When combined with modern modification on transformers, it can further improve the performance, showing its complementaryness with main-stream efforts.
%
%


\subsection{Qualitative Result}
\begin{figure}[t]
\centering
\includegraphics[width=\linewidth]{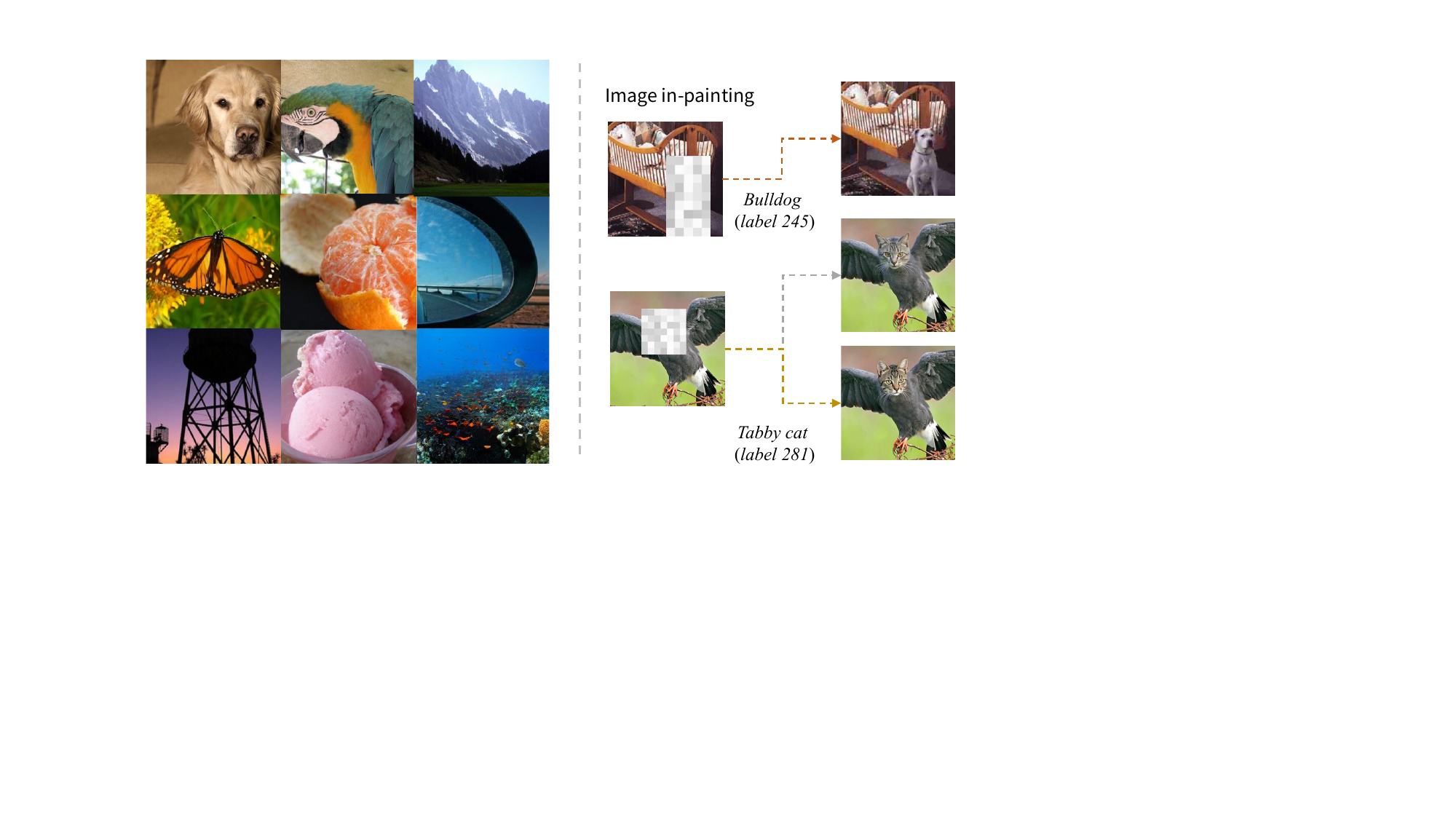}
\caption{Class-conditional generation and in-painting samples of ReDDiT on ImageNet $256\times 256$.}
\label{fig:samples}
\end{figure}

\paragraph{Class-conditional Generation.}
Figure~\ref{fig:samples} presents representative class-conditional samples generated by the proposed ReDDiT model. The outputs across diverse image classes consistently exhibit high fidelity and diversity.
Additional qualitative comparisons and more sample visualizations are provided in Appendix~\ref{appen-quality}.
\paragraph{Image Editing.}
We further demonstrate ReDDiT’s editing capability in Figure~\ref{fig:samples}, highlighting its bi-directional perceptual competence.
Following MaskGIT~\cite{changMaskGITMaskedGenerative2022}, we replace a region of the input image with noise tokens and employ the same generation pipeline to inpaint the missing content, conditioned on a class label $c$.
Thanks to the rehashing noise mechanism, ReDDiT is able to produce diverse and semantically coherent completions without adjusting temperature or other sampling parameters.

\section{Conclusion}
\label{sec:con}

We proposed ReDDiT, a discrete visual generative model built upon a discrete diffusion architecture with novel noise designs and efficient sampling strategies.
Our key contribution lies in the integration of rehashing noise with samplers, which together ensure both diversity and low discrepancy throughout the generative process.
By introducing rehashing noise, ReDDiT enriches the potential paths that latent variables can traverse during training, regularize training dynamics and enhances model's representational capacity. 
Extensive experiments demonstrate that discrete generative models can achieve performance on par with their continuous counterparts while offering top-tier efficiency.
This study paves a promising way for discrete generative modeling and offers fresh insights toward unifying visual and language generation—a path we leave for future exploration.
%
\bibliography{main}

\begin{thebibliography}{52}
\providecommand{\natexlab}[1]{#1}
\providecommand{\url}[1]{\texttt{#1}}
\expandafter\ifx\csname urlstyle\endcsname\relax
  \providecommand{\doi}[1]{doi: #1}\else
  \providecommand{\doi}{doi: \begingroup \urlstyle{rm}\Url}\fi

\bibitem[Austin et~al.(2021)Austin, Johnson, Ho, Tarlow, and van~den Berg]{austinStructuredDenoisingDiffusion2023}
Jacob Austin, Daniel~D. Johnson, Jonathan Ho, Daniel Tarlow, and Rianne van~den Berg.
\newblock Structured denoising diffusion models in discrete state-spaces.
\newblock In \emph{NeurIPS}, pp.\  17981--17993, 2021.

\bibitem[Bai et~al.(2025)Bai, Ye, Chow, Song, Chen, Li, Dong, Zhu, and Yan]{baiMeissonicRevitalizingMasked2025}
Jinbin Bai, Tian Ye, Wei Chow, Enxin Song, Qing-Guo Chen, Xiangtai Li, Zhen Dong, Lei Zhu, and Shuicheng Yan.
\newblock Meissonic: Revitalizing masked generative transformers for efficient high-resolution text-to-image synthesis.
\newblock In \emph{ICLR}, 2025.

\bibitem[Campbell et~al.(2022)Campbell, Benton, De~Bortoli, Rainforth, Deligiannidis, and Doucet]{campbellContinuousTimeFramework2022}
Andrew Campbell, Joe Benton, Valentin De~Bortoli, Thomas Rainforth, George Deligiannidis, and Arnaud Doucet.
\newblock A continuous time framework for discrete denoising models.
\newblock \emph{NeurIPS}, 35:\penalty0 28266--28279, 2022.

\bibitem[Chang et~al.(2022)Chang, Zhang, Jiang, Liu, and Freeman]{changMaskGITMaskedGenerative2022}
Huiwen Chang, Han Zhang, Lu~Jiang, Ce~Liu, and William~T Freeman.
\newblock Maskgit: Masked generative image transformer.
\newblock In \emph{CVPR}, pp.\  11315--11325, 2022.

\bibitem[Chang et~al.(2023)Chang, Zhang, Barber, Maschinot, Lezama, Jiang, Yang, Murphy, Freeman, Rubinstein, et~al.]{changMuseTextImageGeneration2023}
Huiwen Chang, Han Zhang, Jarred Barber, AJ~Maschinot, Jose Lezama, Lu~Jiang, Ming-Hsuan Yang, Kevin Murphy, William~T Freeman, Michael Rubinstein, et~al.
\newblock Muse: Text-to-image generation via masked generative transformers.
\newblock \emph{arXiv preprint arXiv:2301.00704}, 2023.

\bibitem[Deng et~al.(2009)Deng, Dong, Socher, Li, Li, and Fei{-}Fei]{imageNet2009}
Jia Deng, Wei Dong, Richard Socher, Li{-}Jia Li, Kai Li, and Li~Fei{-}Fei.
\newblock Imagenet: {A} large-scale hierarchical image database.
\newblock In \emph{IEEE CVPR}, pp.\  248--255, 2009.

\bibitem[Devlin et~al.(2019)Devlin, Chang, Lee, and Toutanova]{devlinBERTPretrainingDeep2019}
Jacob Devlin, Ming-Wei Chang, Kenton Lee, and Kristina Toutanova.
\newblock Bert: Pre-training of deep bidirectional transformers for language understanding.
\newblock In \emph{NAACL}, pp.\  4171--4186, 2019.

\bibitem[Dhariwal \& Nichol(2021)Dhariwal and Nichol]{dhariwalDiffusionModelsBeat2021}
Prafulla Dhariwal and Alexander Nichol.
\newblock Diffusion models beat gans on image synthesis.
\newblock \emph{NeurIPS}, 34:\penalty0 8780--8794, 2021.

\bibitem[Esser et~al.(2021)Esser, Rombach, and Ommer]{TamingVQGAN}
Patrick Esser, Robin Rombach, and Bj{\"{o}}rn Ommer.
\newblock Taming transformers for high-resolution image synthesis.
\newblock In \emph{CVPR}, pp.\  12873--12883, 2021.

\bibitem[Gao et~al.(2023)Gao, Zhou, Cheng, and Yan]{gaoMDTv2MaskedDiffusion2024}
Shanghua Gao, Pan Zhou, Ming-Ming Cheng, and Shuicheng Yan.
\newblock Mdtv2: Masked diffusion transformer is a strong image synthesizer.
\newblock \emph{arXiv preprint arXiv:2303.14389}, 2023.

\bibitem[Gat et~al.(2024)Gat, Remez, Shaul, Kreuk, Chen, Synnaeve, Adi, and Lipman]{gatDiscreteFlowMatching2024}
Itai Gat, Tal Remez, Neta Shaul, Felix Kreuk, Ricky~TQ Chen, Gabriel Synnaeve, Yossi Adi, and Yaron Lipman.
\newblock Discrete flow matching.
\newblock \emph{Advances in Neural Information Processing Systems}, 37:\penalty0 133345--133385, 2024.

\bibitem[Gu et~al.(2022)Gu, Chen, Bao, Wen, Zhang, Chen, Yuan, and Guo]{guVectorQuantizedDiffusion2022}
Shuyang Gu, Dong Chen, Jianmin Bao, Fang Wen, Bo~Zhang, Dongdong Chen, Lu~Yuan, and Baining Guo.
\newblock Vector quantized diffusion model for text-to-image synthesis.
\newblock In \emph{Proceedings of the IEEE/CVF conference on computer vision and pattern recognition}, pp.\  10696--10706, 2022.

\bibitem[Heusel et~al.(2017)Heusel, Ramsauer, Unterthiner, Nessler, and Hochreiter]{heusel2017gans}
Martin Heusel, Hubert Ramsauer, Thomas Unterthiner, Bernhard Nessler, and Sepp Hochreiter.
\newblock Gans trained by a two time-scale update rule converge to a local nash equilibrium.
\newblock In \emph{NeurIPS}, pp.\  6626--6637, 2017.

\bibitem[Ho \& Salimans(2022)Ho and Salimans]{hoClassifierFreeDiffusionGuidance2022}
Jonathan Ho and Tim Salimans.
\newblock Classifier-free diffusion guidance.
\newblock \emph{arXiv preprint arXiv:2207.12598}, 2022.

\bibitem[Ho et~al.(2020)Ho, Jain, and Abbeel]{hoDenoisingDiffusionProbabilistic2020}
Jonathan Ho, Ajay Jain, and Pieter Abbeel.
\newblock Denoising diffusion probabilistic models.
\newblock \emph{Advances in neural information processing systems}, 33:\penalty0 6840--6851, 2020.

\bibitem[Hu \& Ommer(2024)Hu and Ommer]{huMASKAllYou2024}
Vincent~Tao Hu and Bj{\"o}rn Ommer.
\newblock [mask] is all you need.
\newblock \emph{arXiv preprint arXiv:2412.06787}, 2024.

\bibitem[Huang et~al.(2023)Huang, Mao, Chen, and Zhang]{RQTransformer23}
Mengqi Huang, Zhendong Mao, Zhuowei Chen, and Yongdong Zhang.
\newblock Towards accurate image coding: Improved autoregressive image generation with dynamic vector quantization.
\newblock In \emph{CVPR}, pp.\  22596--22605, 2023.

\bibitem[Lezama et~al.(2022)Lezama, Chang, Jiang, and Essa]{lezamaImprovedMaskedcritic2022}
Jos{\'e} Lezama, Huiwen Chang, Lu~Jiang, and Irfan Essa.
\newblock Improved masked image generation with token-critic.
\newblock In \emph{ECCV}, pp.\  70--86. Springer, 2022.

\bibitem[Li et~al.(2024)Li, Zhang, Lin, Xiong, Long, Deng, Zhang, Liu, Huang, Xiao, et~al.]{liHunyuanDiTPowerfulMultiResolution2024}
Zhimin Li, Jianwei Zhang, Qin Lin, Jiangfeng Xiong, Yanxin Long, Xinchi Deng, Yingfang Zhang, Xingchao Liu, Minbin Huang, Zedong Xiao, et~al.
\newblock Hunyuan-dit: A powerful multi-resolution diffusion transformer with fine-grained chinese understanding.
\newblock \emph{arXiv preprint arXiv:2405.08748}, 2024.

\bibitem[Liu et~al.(2025)Liu, Yang, Zhang, Chen, Zou, Wei, Wang, and Zhang]{liuDLLMCacheAcceleratingDiffusion2025}
Zhiyuan Liu, Yicun Yang, Yaojie Zhang, Junjie Chen, Chang Zou, Qingyuan Wei, Shaobo Wang, and Linfeng Zhang.
\newblock {{dLLM-Cache}}: {{Accelerating Diffusion Large Language Models}} with {{Adaptive Caching}}.
\newblock \emph{arXiv preprint arXiv:2506.06295}, 2025.

\bibitem[Lu et~al.(2022)Lu, Zhou, Bao, Chen, Li, and Zhu]{luDPMSolverFastSolver2023}
Cheng Lu, Yuhao Zhou, Fan Bao, Jianfei Chen, Chongxuan Li, and Jun Zhu.
\newblock Dpm-solver++: Fast solver for guided sampling of diffusion probabilistic models.
\newblock \emph{arXiv preprint arXiv:2211.01095}, 2022.

\bibitem[Ma et~al.(2024)Ma, Goldstein, Albergo, Boffi, {Vanden-Eijnden}, and Xie]{maSiTExploringFlow2024a}
Nanye Ma, Mark Goldstein, Michael~S. Albergo, Nicholas~M. Boffi, Eric {Vanden-Eijnden}, and Saining Xie.
\newblock \emph{{{SiT}}: {{Exploring Flow}} and~{{Diffusion-Based Generative Models}} with~{{Scalable Interpolant Transformers}}}, pp.\  23--40.
\newblock Springer Nature Switzerland, 2024.

\bibitem[Nie et~al.(2025)Nie, Zhu, You, Zhang, Ou, Hu, Zhou, Lin, Wen, and Li]{nieLargeLanguageDiffusion2025}
Shen Nie, Fengqi Zhu, Zebin You, Xiaolu Zhang, Jingyang Ou, Jun Hu, Jun Zhou, Yankai Lin, Ji-Rong Wen, and Chongxuan Li.
\newblock Large language diffusion models.
\newblock \emph{arXiv preprint arXiv:2502.09992}, 2025.

\bibitem[Nisonoff et~al.(2024)Nisonoff, Xiong, Allenspach, and Listgarten]{nisonoffUnlockingGuidanceDiscrete2024}
Hunter Nisonoff, Junhao Xiong, Stephan Allenspach, and Jennifer Listgarten.
\newblock Unlocking guidance for discrete state-space diffusion and flow models.
\newblock \emph{arXiv preprint arXiv:2406.01572}, 2024.

\bibitem[Oquab et~al.(2023)Oquab, Darcet, Moutakanni, Vo, Szafraniec, Khalidov, Fernandez, Haziza, Massa, El-Nouby, et~al.]{oquabDINOv2LearningRobust2024}
Maxime Oquab, Timoth{\'e}e Darcet, Th{\'e}o Moutakanni, Huy Vo, Marc Szafraniec, Vasil Khalidov, Pierre Fernandez, Daniel Haziza, Francisco Massa, Alaaeldin El-Nouby, et~al.
\newblock Dinov2: Learning robust visual features without supervision.
\newblock \emph{arXiv preprint arXiv:2304.07193}, 2023.

\bibitem[Ou et~al.(2024)Ou, Nie, Xue, Zhu, Sun, Li, and Li]{ouYourAbsorbingDiscrete2025}
Jingyang Ou, Shen Nie, Kaiwen Xue, Fengqi Zhu, Jiacheng Sun, Zhenguo Li, and Chongxuan Li.
\newblock Your absorbing discrete diffusion secretly models the conditional distributions of clean data.
\newblock \emph{arXiv preprint arXiv:2406.03736}, 2024.

\bibitem[Pang et~al.(2024)Pang, Zhang, Luan, Man, Tan, Zhang, Freeman, and Wang]{pangRandARDecoderonlyAutoregressive2024}
Ziqi Pang, Tianyuan Zhang, Fujun Luan, Yunze Man, Hao Tan, Kai Zhang, William~T Freeman, and Yu-Xiong Wang.
\newblock Randar: Decoder-only autoregressive visual generation in random orders.
\newblock \emph{arXiv preprint arXiv:2412.01827}, 2024.

\bibitem[Peebles \& Xie(2023)Peebles and Xie]{peeblesScalableDiffusionModels2023}
William Peebles and Saining Xie.
\newblock Scalable diffusion models with transformers.
\newblock In \emph{CVPR}, pp.\  4195--4205, 2023.

\bibitem[Rombach et~al.(2022{\natexlab{a}})Rombach, Blattmann, Lorenz, Esser, and Ommer]{LDM4}
Robin Rombach, Andreas Blattmann, Dominik Lorenz, Patrick Esser, and Bj{\"{o}}rn Ommer.
\newblock High-resolution image synthesis with latent diffusion models.
\newblock In \emph{CVPR}, pp.\  10674--10685, 2022{\natexlab{a}}.

\bibitem[Rombach et~al.(2022{\natexlab{b}})Rombach, Blattmann, Lorenz, Esser, and Ommer]{rombachHighResolutionImageSynthesis2022a}
Robin Rombach, Andreas Blattmann, Dominik Lorenz, Patrick Esser, and Bj{\"o}rn Ommer.
\newblock High-resolution image synthesis with latent diffusion models.
\newblock In \emph{CVPR}, pp.\  10684--10695, 2022{\natexlab{b}}.

\bibitem[Sahoo et~al.(2024)Sahoo, Arriola, Schiff, Gokaslan, Marroquin, Chiu, Rush, and Kuleshov]{sahooSimpleEffectiveMasked2024}
Subham Sahoo, Marianne Arriola, Yair Schiff, Aaron Gokaslan, Edgar Marroquin, Justin Chiu, Alexander Rush, and Volodymyr Kuleshov.
\newblock Simple and effective masked diffusion language models.
\newblock \emph{NeurIPS}, 37:\penalty0 130136--130184, 2024.

\bibitem[Salimans et~al.(2017)Salimans, Goodfellow, Zaremba, Cheung, Radford, and Chen]{IS2016}
Tim Salimans, Ian Goodfellow, Wojciech Zaremba, Vicki Cheung, Alec Radford, and Xi~Chen.
\newblock Improved techniques for training gans.
\newblock In \emph{NeurIPS}, pp.\  2226--2234, 2017.

\bibitem[Santos et~al.(2023)Santos, Fox, Lubbers, and Lin]{santosBlackoutDiffusionGenerative2023}
Javier~E Santos, Zachary~R Fox, Nicholas Lubbers, and Yen~Ting Lin.
\newblock Blackout diffusion: generative diffusion models in discrete-state spaces.
\newblock In \emph{ICML}, pp.\  9034--9059. PMLR, 2023.

\bibitem[Shaul et~al.(2024)Shaul, Gat, Havasi, Severo, Sriram, Holderrieth, Karrer, Lipman, and Chen]{shaulFlowMatchingGeneral2024}
Neta Shaul, Itai Gat, Marton Havasi, Daniel Severo, Anuroop Sriram, Peter Holderrieth, Brian Karrer, Yaron Lipman, and Ricky~TQ Chen.
\newblock Flow matching with general discrete paths: A kinetic-optimal perspective.
\newblock \emph{arXiv preprint arXiv:2412.03487}, 2024.

\bibitem[Shi et~al.(2025)Shi, Luo, Ge, Yang, Shan, and Wang]{shiScalableImageIBQ2025}
Fengyuan Shi, Zhuoyan Luo, Yixiao Ge, Yujiu Yang, Ying Shan, and Limin Wang.
\newblock Scalable {{Image Tokenization}} with {{Index Backpropagation Quantization}}.
\newblock \emph{arXiv preprint arXiv:2412.02692}, 2025.

\bibitem[Shi et~al.(2024)Shi, Han, Wang, Doucet, and Titsias]{shiSimplifiedGeneralizedMasked2025}
Jiaxin Shi, Kehang Han, Zhe Wang, Arnaud Doucet, and Michalis Titsias.
\newblock Simplified and generalized masked diffusion for discrete data.
\newblock \emph{NeurIPS}, 37:\penalty0 103131--103167, 2024.

\bibitem[Song et~al.(2020)Song, Meng, and Ermon]{songDenoisingDiffusionImplicit2022}
Jiaming Song, Chenlin Meng, and Stefano Ermon.
\newblock Denoising diffusion implicit models.
\newblock \emph{arXiv preprint arXiv:2010.02502}, 2020.

\bibitem[Su et~al.(2024)Su, Ahmed, Lu, Pan, Bo, and Liu]{suRopeFormerEnhancedTransformer2023}
Jianlin Su, Murtadha Ahmed, Yu~Lu, Shengfeng Pan, Wen Bo, and Yunfeng Liu.
\newblock Roformer: Enhanced transformer with rotary position embedding.
\newblock \emph{Neurocomputing}, 568:\penalty0 127063, 2024.

\bibitem[Sun et~al.(2024)Sun, Jiang, Chen, Zhang, Peng, Luo, and Yuan]{sunllamagen2024}
Peize Sun, Yi~Jiang, Shoufa Chen, Shilong Zhang, Bingyue Peng, Ping Luo, and Zehuan Yuan.
\newblock Autoregressive model beats diffusion: Llama for scalable image generation.
\newblock \emph{arXiv preprint arXiv:2406.06525}, 2024.

\bibitem[Swerdlow et~al.(2025)Swerdlow, Prabhudesai, Gandhi, Pathak, and Fragkiadaki]{swerdlowUnifiedMultimodalDiscrete2025}
Alexander Swerdlow, Mihir Prabhudesai, Siddharth Gandhi, Deepak Pathak, and Katerina Fragkiadaki.
\newblock Unified multimodal discrete diffusion.
\newblock \emph{arXiv preprint arXiv:2503.20853}, 2025.

\bibitem[Tian et~al.(2024)Tian, Jiang, Yuan, Peng, and Wang]{tian_VAR_2024}
Keyu Tian, Yi~Jiang, Zehuan Yuan, Bingyue Peng, and Liwei Wang.
\newblock Visual autoregressive modeling: Scalable image generation via next-scale prediction.
\newblock \emph{NeurIPS}, 37:\penalty0 84839--84865, 2024.

\bibitem[Wang et~al.(2025)Wang, Li, {zhang}, Song, Li, Ge, Zheng, and Wang]{wangDifferentiableSolverSearch2025}
Shuai Wang, Zexian Li, Qipeng {zhang}, Tianhui Song, Xubin Li, Tiezheng Ge, Bo~Zheng, and Limin Wang.
\newblock Differentiable {{Solver Search}} for {{Fast Diffusion Sampling}}.
\newblock \emph{arXiv preprint arXiv:2505.21114}, 2025.

\bibitem[Wu et~al.(2024)Wu, Hu, and Wei]{wuRDPMSolveDiffusion2024}
Xiaoping Wu, Jie Hu, and Xiaoming Wei.
\newblock Rdpm: Solve diffusion probabilistic models via recurrent token prediction.
\newblock \emph{arXiv preprint arXiv:2412.18390}, 2024.

\bibitem[Xin et~al.(2025)Xin, Zhuo, Qin, Luo, Cao, Fu, He, Li, Zhai, Liu, and Gao]{xinResurrectMaskAutoRegressive2025}
Yi~Xin, Le~Zhuo, Qi~Qin, Siqi Luo, Yuewen Cao, Bin Fu, Yangfan He, Hongsheng Li, Guangtao Zhai, Xiaohong Liu, and Peng Gao.
\newblock Resurrect {{Mask AutoRegressive Modeling}} for {{Efficient}} and {{Scalable Image Generation}}.
\newblock \emph{arXiv preprint arXiv:2507.13032}, 2025.

\bibitem[Yang et~al.(2025)Yang, Tian, Li, Zhang, Shen, Tong, and Wang]{mmada2025}
Ling Yang, Ye~Tian, Bowen Li, Xinchen Zhang, Ke~Shen, Yunhai Tong, and Mengdi Wang.
\newblock Mmada: Multimodal large diffusion language models.
\newblock \emph{arXiv preprint arXiv:2505.15809}, 2025.

\bibitem[Yu et~al.(2022)Yu, Li, Koh, Zhang, Pang, Qin, Ku, Xu, Baldridge, and Wu]{ViTVQGAN22}
Jiahui Yu, Xin Li, Jing~Yu Koh, Han Zhang, Ruoming Pang, James Qin, Alexander Ku, Yuanzhong Xu, Jason Baldridge, and Yonghui Wu.
\newblock Vector-quantized image modeling with improved vqgan.
\newblock In \emph{ICLR}, 2022.

\bibitem[Yu et~al.(2023)Yu, Lezama, Gundavarapu, Versari, Sohn, Minnen, Cheng, Birodkar, Gupta, Gu, et~al.]{yuLanguageModelBeats2024}
Lijun Yu, Jos{\'e} Lezama, Nitesh~B Gundavarapu, Luca Versari, Kihyuk Sohn, David Minnen, Yong Cheng, Vighnesh Birodkar, Agrim Gupta, Xiuye Gu, et~al.
\newblock Language model beats diffusion--tokenizer is key to visual generation.
\newblock \emph{arXiv preprint arXiv:2310.05737}, 2023.

\bibitem[Yu et~al.(2024)Yu, Weber, Deng, Shen, Cremers, and Chen]{yutitok2024}
Qihang Yu, Mark Weber, Xueqing Deng, Xiaohui Shen, Daniel Cremers, and Liang-Chieh Chen.
\newblock An image is worth 32 tokens for reconstruction and generation.
\newblock \emph{Advances in Neural Information Processing Systems}, 37:\penalty0 128940--128966, 2024.

\bibitem[Yu et~al.(2025)Yu, Kwak, Jang, Jeong, Huang, Shin, and Xie]{RePA24}
Sihyun Yu, Sangkyung Kwak, Huiwon Jang, Jongheon Jeong, Jonathan Huang, Jinwoo Shin, and Saining Xie.
\newblock Representation alignment for generation: Training diffusion transformers is easier than you think.
\newblock In \emph{ICLR}, 2025.

\bibitem[Zhang \& Sennrich(2019)Zhang and Sennrich]{hangsnrEfficientDiffusion2023}
Biao Zhang and Rico Sennrich.
\newblock Root mean square layer normalization.
\newblock \emph{NeurIPS}, 32, 2019.

\bibitem[Zhang et~al.(2023)Zhang, Rao, and Agrawala]{zhangAddingConditionalControl2023}
Lvmin Zhang, Anyi Rao, and Maneesh Agrawala.
\newblock Adding conditional control to text-to-image diffusion models.
\newblock In \emph{ICCV}, pp.\  3836--3847, 2023.

\bibitem[Zheng et~al.(2024)Zheng, Chen, Mao, Liu, Zhu, and Zhang]{zhengMaskedDiffusionModels2025}
Kaiwen Zheng, Yongxin Chen, Hanzi Mao, Ming-Yu Liu, Jun Zhu, and Qinsheng Zhang.
\newblock Masked diffusion models are secretly time-agnostic masked models and exploit inaccurate categorical sampling.
\newblock \emph{arXiv preprint arXiv:2409.02908}, 2024.

\end{thebibliography}
\bibliographystyle{ICLR_2026_style/iclr2026_conference}

\appendix

\section{Use of Large Language Models}
In the process of drafting this paper, Large Language Models (LLMs) were solely utilized for \textbf{writing polish} (\textit{e.g.,} optimizing sentence structure, enhancing expression fluency). No LLM was involved in core academic work such as conceptualization, literature review, data analysis, argument construction, or conclusion formulation of this study.

As the human authors of this paper, we bear full and sole responsibility for the paper’s content, including the accuracy of research data, validity of academic arguments, integrity of research methods, and compliance with academic ethics.
\section{Related Work}
\label{sec:related}

\paragraph{Diffusion Models.}
Diffusion models~\cite{hoDenoisingDiffusionProbabilistic2020, songDenoisingDiffusionImplicit2022} have emerged as a powerful class of generative methods that learn data distributions by reversing a gradual noising process over time.
These models are primarily designed for continuous domains such as images \cite{dhariwalDiffusionModelsBeat2021,gaoMDTv2MaskedDiffusion2024,peeblesScalableDiffusionModels2023}, defining a forward process that transforms data \( x_0 \) into noise \( x_1 \):
$x_t \sim \mathcal{N}(\sqrt{\alpha_t}x_0; (1 - \alpha_t)\mathbf{I})$
where $\alpha_t$ controls the noise schedule. 
The generative (reverse) process learns a denoising model $p_\theta(x_s \mid x_t)$, often parameterized via a neural network $\theta$ to predict either noise or clean data.

\paragraph{Discrete Diffusion Models.}
Discrete diffusion has been previously governed by masked visual token models (MVTMs)~\cite{changMaskGITMaskedGenerative2022,changMuseTextImageGeneration2023, guVectorQuantizedDiffusion2022, yuLanguageModelBeats2024, yutitok2024}.
This model leverages a BERT-style \texttt{[mask]} token to corrupt the tokenized image sequence and trained the network with a simple cross-entropy loss on masked tokens, resulting in a score-based prediction.
It generates tokens in a non-autoregressive fashion, by remasking the tokens with least scores at each inference as depicted in Alg.~\ref{algo-mvtm}.

Recent studies unlocked the principled discrete diffusion model (DDM)~\cite{sahooSimpleEffectiveMasked2024,shiSimplifiedGeneralizedMasked2025} and discrete flow-matching (DFM)~\cite{gatDiscreteFlowMatching2024, shaulFlowMatchingGeneral2024}, which adapt the Markov chain theory, enabling generation over text~\cite{ouYourAbsorbingDiscrete2025, nieLargeLanguageDiffusion2025}, molecules\cite{shaulFlowMatchingGeneral2024}, and other discrete representations~\cite{austinStructuredDenoisingDiffusion2023, nisonoffUnlockingGuidanceDiscrete2024}.
Unlike MVTMs, the principled DDM and DFM mostly derive a time-weighted cross-entropy loss to supervise the training procedure and apply a gradual unmasking method based on probabilities.

\section{Discrete Diffusion with Rehashing Noise}\label{proof-appen}
\paragraph{Complete Definition and Deduction.}
We provide a full theoretical discussion on the corrupted distribution and reverse process defined in the main paper. 
The extended discussions with corresponding proofs are marked with \textcolor{teal}{\textbf{teal}}.

Given $d$ categories, let $\mathbf{e}_i\in \mathbb{R}^{d}$ be its one-hot vector where the $i$-th value is $1$. 
We denote $ \mathcal{E}=\{ \mathbf{e}_i \in \mathbb{R}^{d} \mid i = 1, \ldots, d \}$ as the basis of a categorical distribution, 
and a basis for absorbing states with capacity $m$:  $\mathcal{M}=\{ \mathbf{m}_j \in \mathbb{R}^{m}\mid j = 1, \ldots, m \}$. 
The sum of $\mathcal{E}$ and $\mathcal{M}$ can be denoted as 
\begin{equation}
\mathcal{V}_{(d,m)} \triangleq \left\{ \mathbf{v}_{(i,j)} \in \mathbb{R}^{d+m} \,\middle|\, 
\mathbf{v}_{(i,j)} = 
\begin{cases}
\mathbf{e}_i \oplus \mathbf{0}_m, & \text{for } i = 1, \ldots, d,\ j=0 \\
\mathbf{0}_d \oplus \mathbf{m}_j, & \text{for } j = 1, \ldots, m, \ i=0
\end{cases}
\right\}.
\end{equation}
We further denote the subspace $\mathcal{E}_d, \ \mathcal{M}_m\in \mathcal{V}_{(d,m)}$ which contain either valid or mask tokens, as
\begin{equation}
\mathcal{E}_d = \left\{ \mathbf{v}_{(i,0)} \in \mathcal{V}_{(d,m)} \,\middle|\, i = 1, \ldots, d \right\}, \ \mathcal{M}_m = \left\{ \mathbf{v}_{(0,j)} \in \mathcal{V}_{(d,m)} \,\middle|\, j = 1, \ldots, m \right\}.
\end{equation}
To exploit visits across all the possible paths, for $0\leq s<t\leq 1$, we write the transition kernel as\footnote{To maintain simplicity, we use $\alpha_{t|s}^{\leftarrow}=\frac{\alpha_t}{\alpha_s}$ and $\alpha_{t|s}^{\rightarrow}=\frac{1-\alpha_s}{1-\alpha_t}$ to denote transition rate for the corruption and reverse process, respectively.}
\begin{equation}\label{rewrite-appen}
    q(x_t^i \mid x_s^i) = 
    \begin{cases}
    1 - \alpha_{t|s}^{\leftarrow}, & \text{if } x_t^i \in \mathcal{M}_m,\ x_s^i\notin\mathcal{M}_m, \\
    \alpha_{t|s}^{\leftarrow}, & \text{if } x_t^i = x_s^i,\ x_s^i\notin\mathcal{M}_m, \\
    1/m, & \text{if } x_t^i \in\mathcal{M}_m, \ x_s^i\in\mathcal{M}_m, \\
    0, & \text{otherwise}.
    \end{cases}
\end{equation}

\paragraph{\color{teal}Proof of the Corrupted Distribution.}
The presentation in the main paper simplifies the theory without specifying the transition matrix $Q_t$ due to page limitation. We make a detailed version with important yet basic matrix calculation in this section.

Let \( \mathbf I_{(d, m)} \), \( \mathbf M_{(d, m)} \) and \( \bm\pi _{(d, m)}\) be matrices in \( \mathbb{R}^{(d + m) \times (d + m)} \), defined as

\begin{equation}
\mathbf I_{(d, m)} = 
\begin{bmatrix}
I_d & 0 \\
0 & 0
\end{bmatrix},\quad
\mathbf M_{(d, m)} = 
\begin{bmatrix}
0 & \frac{1}{m} \mathbf{1}_{d \times m} \\
0 & 0
\end{bmatrix}, \quad
\bm \pi_{(d, m)} = 
\begin{bmatrix}
0 & 0 \\
0 & \frac{1}{m} \mathbf{1}_m \mathbf{1}_m^\top
\end{bmatrix} 
\end{equation}

where \( \mathbf I_d \) is the \( d \times d \) identity matrix, and \( \mathbf{1}_m \in \mathbb{R}^m \) is a vector of ones.

The transition matrix \( Q_{t|s} \in \mathbb{R}^{(d + m) \times (d + m)} \) is defined as:
\begin{equation}\label{Qmatrix}
Q_{t|s} = \alpha_{t|s}^{\leftarrow} \mathbf I_{(d, m)} + (1 - \alpha_{t|s}^{\leftarrow}) \mathbf M_{(d, m)} +\bm \pi_{(d, m)}
\end{equation}
which can be demonstrated intuitively:
\vspace{1em}
\[
Q_{t|s} =
\begin{bmatrix}
\alpha_{t|s}^{\leftarrow} & 0 & \cdots & 0 & \frac{1-\alpha_{t|s}^{\leftarrow}}{m} & \frac{1-\alpha_{t|s}^{\leftarrow}}{m} & \cdots & \frac{1-\alpha_{t|s}^{\leftarrow}}{m} \\
0 & \alpha_{t|s}^{\leftarrow} & \cdots & 0 & \frac{1-\alpha_{t|s}^{\leftarrow}}{m} & \frac{1-\alpha_{t|s}^{\leftarrow}}{m} & \cdots & \frac{1-\alpha_{t|s}^{\leftarrow}}{m} \\
\vdots & \vdots & \ddots & \vdots & \vdots & \vdots & \ddots & \vdots \\
0 & 0 & \cdots & \alpha_{t|s}^{\leftarrow} & \frac{1-\alpha_{t|s}^{\leftarrow}}{m} & \frac{1-\alpha_{t|s}^{\leftarrow}}{m} & \cdots & \frac{1-\alpha_{t|s}^{\leftarrow}}{m} \\
0 & 0 & \cdots & 0 & \frac{1}{m} & \frac{1}{m} & \cdots & \frac{1}{m} \\
0 & 0 & \cdots & 0 & \frac{1}{m} & \frac{1}{m} & \cdots & \frac{1}{m} \\
\vdots & \vdots & \ddots & \vdots & \vdots & \vdots & \ddots & \vdots \\
0 & 0 & \cdots & 0 & \frac{1}{m} & \frac{1}{m} & \cdots & \frac{1}{m}
\end{bmatrix}
\]
\vspace{-1em}
\[
\quad \quad \quad \ 
\begin{array}{c}
\underbrace{\hspace{8em}}_{\text{\small \( \times d \)}}
\ \ \ \ \ \ \ \ \ 
\underbrace{\hspace{10em}}_{\text{\small \( \times m \)}}
\end{array}
\]

The corrupted data distribution is a direct derivative of Eq.~\ref{Qmatrix} by setting $s=0$:
\begin{flalign}
x_t &= x_0 Q_{t|0} \nonumber \\
    &= \alpha_tx_0\mathbf I_{(d, m)} + (1-\alpha_t)x_0\mathbf M_{(d, m)} + x_0\bm \pi_{(d, m)} \nonumber\\
    & =\alpha_tx_0 + (1-\alpha_t)x_0\mathbf M_{(d, m)}\nonumber \\
    &\sim \alpha_{t} x_0 + (1 - \alpha_{t}) \, \text{U}(\mathcal{M}_m^L)
\end{flalign}
where $\text{U}(\mathcal{M}_m^L)$ is the uniform distribution on $\mathcal{M}_m^L$. 

\paragraph{\color{teal}Proof of the Reverse Process.}
To generate a sequence of length $L$, the reverse process starts with $x_1 \sim \text{U}(\mathcal{M}_m^L)$. Let $\mathbf a\odot\mathbf b$ denote the Hadamard product between two vectors $\mathbf{a}$ and $\mathbf{b}$, the reverse process is inferred as:

\begin{flalign}\label{qst}
q(x_s \mid x_t) &= \dfrac{Q_{t|s} x_t\odot Q_{s|0}^{\top}x_0}{x_t^{\top}Q_{t|0}^{\top}x_0} \ \ \ \ \text{(D3PM deduction)}\nonumber\\
    &= \dfrac{[\alpha_{t|s}^{\leftarrow}\mathbf I_{(d, m)} x_t +(1-\alpha_{t|s}^{\leftarrow})\mathbf M_{(d, m)}x_t+\bm \pi_{(d, m)}x_t]\odot [\alpha_{s}x_0 +(1-\alpha_{s})\mathbf M_{(d, m)}^{\top}x_0]}
    {x_t^{\top}[\alpha_{t} x_0 +(1-\alpha_{t})\mathbf M_{(d, m)}^{\top}x_0+\pi_{(d, m)}^{\top}x_0]}\nonumber\\
    &= \dfrac{[\alpha_{t|s}^{\leftarrow} \mathbf I_{(d, m)}x_t +(1-\alpha_{t|s}^{\leftarrow})\mathbf M_{(d, m)}x_t+\bm \pi_{(d, m)}x_t]\odot [\alpha_{s} x_0 +(1-\alpha_{s})\mathbf M_{(d, m)}^{\top}x_0]}
    {\alpha_{t}x_t^{\top}x_0 +(1-\alpha_{t})x_t^{\top}\mathbf M_{(d, m)}^{\top}x_0}
\end{flalign}

We consider the separate cases: $x_t^i=x_0^i$ and $x_t^i\in\mathcal{M}_m$.
\paragraph{Case 1.}
For $x_t^i=x_0^i$, Eq.~\ref{qst} is simplified as
\begin{flalign}
q(x_s^i \mid x_t^i=x_0^i)    
    &= \dfrac{\alpha_{t|s}^{\leftarrow} x_0^i \odot \alpha_{s} x_0^i }
    {\alpha_{t}x_0^{i\ \top}x_0^i} \nonumber\\
    &=1
\end{flalign}
\paragraph{Case 2.}
For $x_t^i\in\mathcal{M}_m$, we have
\begin{flalign}
q(x_s^i \mid x_t^i\in\mathcal{M}_m)    
    &= \dfrac{[(1-\alpha_{t|s}^{\leftarrow})\mathbf M_{(d, m)}x_t^i+\bm \pi_{(d, m)}x_t^i]\odot [\alpha_{s} x_0 +(1-\alpha_{s})\mathbf M_{(d, m)}^{\top}x_0]}
    {(1-\alpha_{t})x_t^{i \ \top}\mathbf M_{(d, m)}^{\top}x_0}\nonumber\\
    &= \dfrac{[(1-\alpha_{t|s}^{\leftarrow})\alpha_s\mathbf M_{(d, m)}x_t^{i}\odot x_0 + \bm \pi_{(d, m)}(1-\alpha_{s})x_t^{i}\odot\mathbf M_{(d, m)}^{\top}x_0]}
    {(1-\alpha_{t})x_t^{i \ \top}\mathbf M_{(d, m)}^{\top}x_0}\nonumber\\
    &= \dfrac{(\alpha_s-\alpha_{t})\mathbf M_{(d, m)}x_t^{i}\odot x_0 + (1-\alpha_{s})\bm \pi_{(d, m)}x_t^{i}\odot\mathbf M_{(d, m)}^{\top}x_0}
    {(1-\alpha_{t})x_t^{i \ \top}\mathbf M_{(d, m)}^{\top}x_0}
\end{flalign}
Notice that $\alpha_{t|s}^{\rightarrow}=\frac{1-\alpha_s}{1-\alpha_t}$, and we have 
\begin{equation}
q(x_s^i\in \mathcal{M}_m\mid x_t^i\in\mathcal{M}_m)=\dfrac{1-\alpha_s}{m(1-\alpha_t)}=\dfrac{\alpha_{t|s}^{\rightarrow}}{m}
\end{equation}
\begin{equation}
q(x_s^i\notin \mathcal{M}_m\mid x_t^i\in\mathcal{M}_m)=\dfrac{\alpha_s-\alpha_t}{1-\alpha_t}=1-\alpha_{t|s}^{\rightarrow}
\end{equation}
Combining case 1 with case 2, we have
\begin{equation}\label{reverse-appen}
    q(x_s^i \mid x_t^i) = 
    \begin{cases}
    1, & \text{if } x_s^i = x_t^i, \ x_t^i \notin \mathcal{M}_m ,\\
    \alpha_{t|s}^{\rightarrow}/m, & \text{if } x_s^i \in \mathcal{M}_m, \ x_t^i \in \mathcal{M}_m ,\\
    1 - \alpha_{t|s}^{\rightarrow}, & \text{if } x_s^i \notin \mathcal{M}_m, \ x_t^i \in \mathcal{M}_m ,  \\
    0, & \text{otherwise.} \\
    \end{cases}
\end{equation}
Following MDLM's deduction, assume that the denoising network can reconstruct $x_0$ perfectly, we use $p_\theta(x_t)$ to approximate this reverse process for complex sequences, and get
\begin{equation}\label{reverse-appen-1}
   q(x_{s}^i|x_{t})=
    \begin{cases}
    1, & \text{if } x_s^i = x_t^i, \ x_t^i \notin \mathcal{M}_m ,\\
    \alpha_{t|s}^{\rightarrow}/m, & \text{if } x_s^i \in \mathcal{M}_m, \ x_t^i \in \mathcal{M}_m ,\\
    (1 - \alpha_{t|s}^{\rightarrow})p_\theta^i(x_t), & \text{if } x_s^i \notin \mathcal{M}_m, \ x_t^i \in \mathcal{M}_m ,  \\
    0, & \text{otherwise.} \\
    \end{cases}
\end{equation}

\section{Sampling from Learned Networks}\label{appen-dfm}
\begin{figure}[t]
\centering
\includegraphics[width=0.8\linewidth]{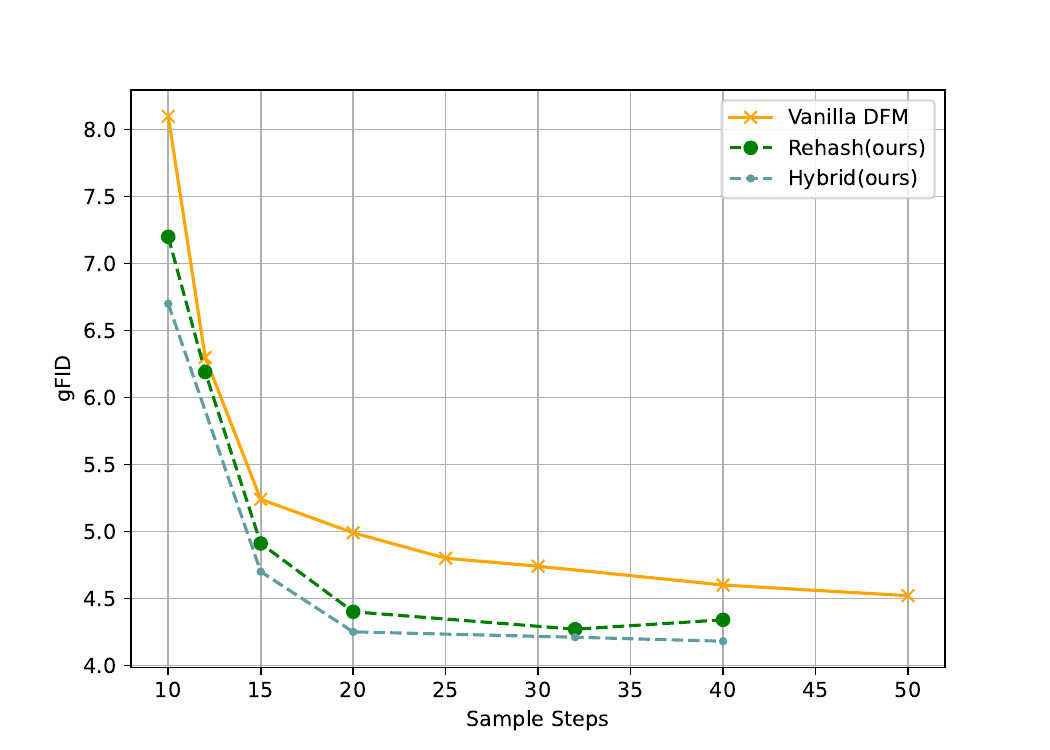}
\caption{Generation quality comparison with DFM methods. The experiments are conducted on ReDDiT-L with a constant classifier-free guidance ($\text{cfg}=2.0$).}
\label{fig:hybrid}
\end{figure}

We present a detailed version of discrete flow matching (DFM) sampler~\ref{dfm} and ours~\ref{rehash}, and discuss the integration of them. Fig.~\ref{fig:hybrid} presents a quantitative comparison of the vanilla DFM sampler, our proposed rehash sampler, and a hybrid strategy that combines both approaches by incorporating selected DFM steps into the rehash trajectory. All methods are evaluated using identical model weights, as the training objectives are compatible due to their shared time-weighted loss formulation.

The rehash sampler exhibits stronger overall performance than DFM, especially in the 15–32 step range, where it achieves low and stable gFID scores. This suggests that our modification enables more efficient decoding trajectories without sacrificing sample quality. The hybrid variant, which integrates only the middle and final steps of the DFM update into the rehash schedule, also delivers consistent gains over the vanilla DFM, suggesting that partial refinement from DFM is beneficial even when the majority of the trajectory is governed by our rehash dynamics.

By leveraging shared gradual decoding infrastructure, the hybrid approach enables practical integration of DFM refinement into the ReDDiT framework with minimal overhead. As noted in the main paper, this leads to a $\sim$0.1 improvement in gFID on ImageNet-1K, reinforcing the complementary strengths of the two samplers. We leave the comprehensive study on the optimal integration of different samplers for future exploration.

\begin{algorithm}[t]
\caption{DFM Sampling Stepwise Pseudo Code}
\label{dfm}
\begin{algorithmic}[1]
\Require $x_t$, labels, timestep $t$, step size $\Delta t$
\State Compute jump probabilities: $j_t \gets 1-\alpha_t$, $j_s \gets 1-\alpha_{t - \Delta t}$
\State Determine guidance scale $\omega$ from schedule
\State Obtain logits $\text{logits}_{\text{cond}}$, $\text{logits}_{\text{uncond}}$ via forward pass
\State $\text{logits}_{x_0} \gets \text{logits}_{\text{uncond}} + \omega \cdot (\text{logits}_{\text{cond}} - \text{logits}_{\text{uncond}})$
\State $p_{x_0} \gets \texttt{softmax}(\text{logits}_{x_0})$
\State Sample $\hat{x}_0 \sim p_{x_0}$ using categorical sampling
\State Construct one-hot encodings: $\delta_{x_0}, \delta_{x_t}$
\State $\text{corrective} \gets \frac{j_s}{j_t} \cdot \delta_{x_t}$
\State $u \gets \frac{j_t - j_s}{j_t} \cdot \delta_{x_0}$
\State Overwrite $u$ in masked range with $\text{corrective}$ terms
\State Mask entries already present in $x_t$ from $u$
\State Compute total transition intensity: $\lambda \gets \sum u$, elementwise
\State Draw Bernoulli mask: $M \sim \text{Bernoulli}(1 - \exp(-\lambda))$
\State For each masked position in $M$, sample from categorical $u$ to obtain updated $x_s$
\State \Return $x_s$
\end{algorithmic}
\end{algorithm}

\begin{algorithm}[t]
\caption{Rehash Sampling (ours) Stepwise Pseudo Code}
\label{rehash}
\begin{algorithmic}[1]
\Require $x_t$, labels, timestep $t$, step size $\Delta t$ (determined by $T^{k+1}-T^{k}$)
\State Indentify the masked tokens: $M \gets [x_t\in \mathcal{M}_m]$ 
\State Rehash $x_t$: $x_t \gets M\cdot\texttt{random\_shuffle}(\text{mask\_vocab}) + (1-M)\cdot x_t$.
\State Compute move coefficients: $k_t \gets 1-\alpha_t$, $k_s \gets 1-\alpha_{t - \Delta t}$
\State Determine guidance scale $\omega$ from schedule
\State Obtain logits $\text{logits}_{\text{cond}}$, $\text{logits}_{\text{uncond}}$ via forward pass
\State $\text{logits}_{\text{cfg}} \gets \text{logits}_{\text{uncond}} + \omega \cdot (\text{logits}_{\text{cond}} - \text{logits}_{\text{uncond}})$
\State $p_{x_0} \gets \texttt{softmax}(\text{logits}_{\text{cfg}})$
\State Set mask probability: $p_{\text{mask}} \gets k_s$
\State Construct proposal distribution: $q_{x_s} \gets \frac{k_t - k_s}{k_t} \cdot p_{x_0}$
\State Overwrite mask token logits:
    \[
    q_{x_s}[:,:,m{:}] \gets 0, \quad
    q_{x_s}[:,:,m] \gets \frac{p_{\text{mask}}}{k_t}
    \]
    where $m$ is the start of mask token index
\State Sample $\hat{x} \sim q_{x_s}$ using categorical sampling
\State Identify preserved tokens: $c \gets [x_t < m]$
\State Combine result: $x_s \gets c \cdot x_t + (1 - c) \cdot \hat{x}$
\State \Return $x_s$
\end{algorithmic}
\end{algorithm}

\section{Experiment Details}\label{appen-details}
We provide detailed training and generation configurations for ReDDiT in Table~\ref{tab:training-details}. Our method incorporates DINOv2-B for representation alignment, which requires computing image features during the forward pass (only activated during training). This introduces an overhead, making training roughly 1.2× slower than solely on discrete tokens. However, this additional cost is offset by faster convergence and improved stability, particularly in early training stages.

The use of quantized latents allows for larger batch sizes under limited GPU memory, making our approach more accessible for low-resource settings. Additionally, aligning discrete codes with semantic features improves the quality and diversity of learned representations. Overall, our design balances computational efficiency with model performance, making it a practical choice for both research and deployment. 

\begin{table}[t]
\centering
\caption{Experiment details for ReDDiT on ImageNet-1K. \textit{Vari.} refers to a time-variant growing guidance scale following MDTv2, which is a common practice for diffusion models.}
\label{tab:training-details}
\scalebox{0.9}{
\begin{tabular}{l|cccc}
\toprule
\textbf{Setting} & \textbf{ReDDiT-L (Ablation)} & \textbf{ReDDiT-L} & \textbf{ReDDiT-XL} & \textbf{ReDDiT-XL$\rm_{f8}$} \\
\midrule
Hidden Size         & 1024  & 1024            & 1280                   & 1280 \\
Transformer Block         & 24  & 24            & 28                   & 28 \\
Attention Head         & 16  & 16            & 20                   & 20 \\
Image Tokenizer            & LlamaGen-f16 & IBQ-f16           & IBQ-f16         & LlamaGen-f8 \\
Codebook Size       & 16384  & 16384            & 16384                  & 16384 \\
Noise Capacity       & 128  & 1024            & 1024                  & 128 \\
Sequence Length     & 256 & 256           & 256                 & 1024 \\
RepA Latent Size       & 16×16   & 16×16            & 16×16                   & 32×32 \\
Batch Size         & 64  & 64            & 32                   & 16 \\
Global Batch Size        & 1024   & 1024            & 1024                   & 1024 \\
LR scheduler     & Cosine Decay   & Cosine Decay           & Cosine Decay                  & Cosine Decay \\
Learning Rate       & 3e-4 & 3e-4           & 3e-4                  & 4e-4 \\
Minimal LR      & 1e-5  & 1e-5           & 1e-5                  & 1e-5 \\
Warmup Steps        & 2k & 2k             & 2k                   & 2k \\
Training Steps      & 500k & 500k           & 500k                  & 500k \\
Training Time      & $\sim$1 day  & $\sim$1 day        & $\sim$2 days               & $\sim$3 days \\
Generation CFG (\textit{Vari.})      & 1.0-5.0 & 1.0-6.5 & 1.0-6.5  & 1.0-5.5 \\
\bottomrule
\end{tabular}
}
\end{table}

\section{Accelerating ReDDiT}\label{acc}
Recent efforts on scaling and accelerating discrete diffusion models are making this generative paradigm more practical than theoretical attempts. We adapt the dLLM-Cache~\cite{liuDLLMCacheAcceleratingDiffusion2025} design into our framework, which efficiently reuses intermediate computations without compromising model performance. Since the condition is modulated using AdaLN and introduces minimal calculation, we do not activate $K_p$ (cache for prompt). As the decoding of visual sequence varies with time more quickly than language decoding, we implement the cache for response with small values like $K_r=2 \ \text{or}\ 4$, which means the $K$ and $V$ of transformer layer is updated every $2$ or $4$ decoding steps instead of per step. As shown in Tab.~\ref{table:acc}, the inference speed is boosted up to 2 times with minimal performance drop, which makes our largest model ReDDiT-XL$\rm _{f8}$ comparable to diffusion models with accelerated solvers.
\begin{table}[t]
\caption{\textbf{Acceleration of ReDDiT using response cache $K_r$.}}
\label{table:acc}
\centering
\scalebox{1.0}{
\begin{tabular}{lccccc}
\toprule
\multirow{2}{*}{Model} & \multicolumn{2}{c}{Config} & \multicolumn{2}{c}{Performance} \\
\cmidrule(lr){2-3} \cmidrule(lr){4-5}
 & Steps & $K_r$ & Relative Speed & gFID$\downarrow$ \\
\midrule
ReDDiT-L &\multirow{2}{*}{32} & \multirow{2}{*}{2} & $\times 1.33$ & 2.28 ($\Delta=0.15$)   \\
ReDDiT-XL &  &  & $\times 1.52$ & 1.88 ($\Delta=0.14$)  \\
\midrule
ReDDiT-XL & \multirow{2}{*}{64} & \multirow{2}{*}{4} & $\times 2.17$ & 1.83 ($\Delta=0.09$)  \\
ReDDiT-XL$\rm _{f8}$  &  &  & $\times 2.56$ & 1.71 ($\Delta=0.10$) \\
\bottomrule
\end{tabular}
}
\end{table}

\section{Qualitative Results}\label{appen-quality}
\begin{figure}[t]
\centering
\includegraphics[width=\linewidth]{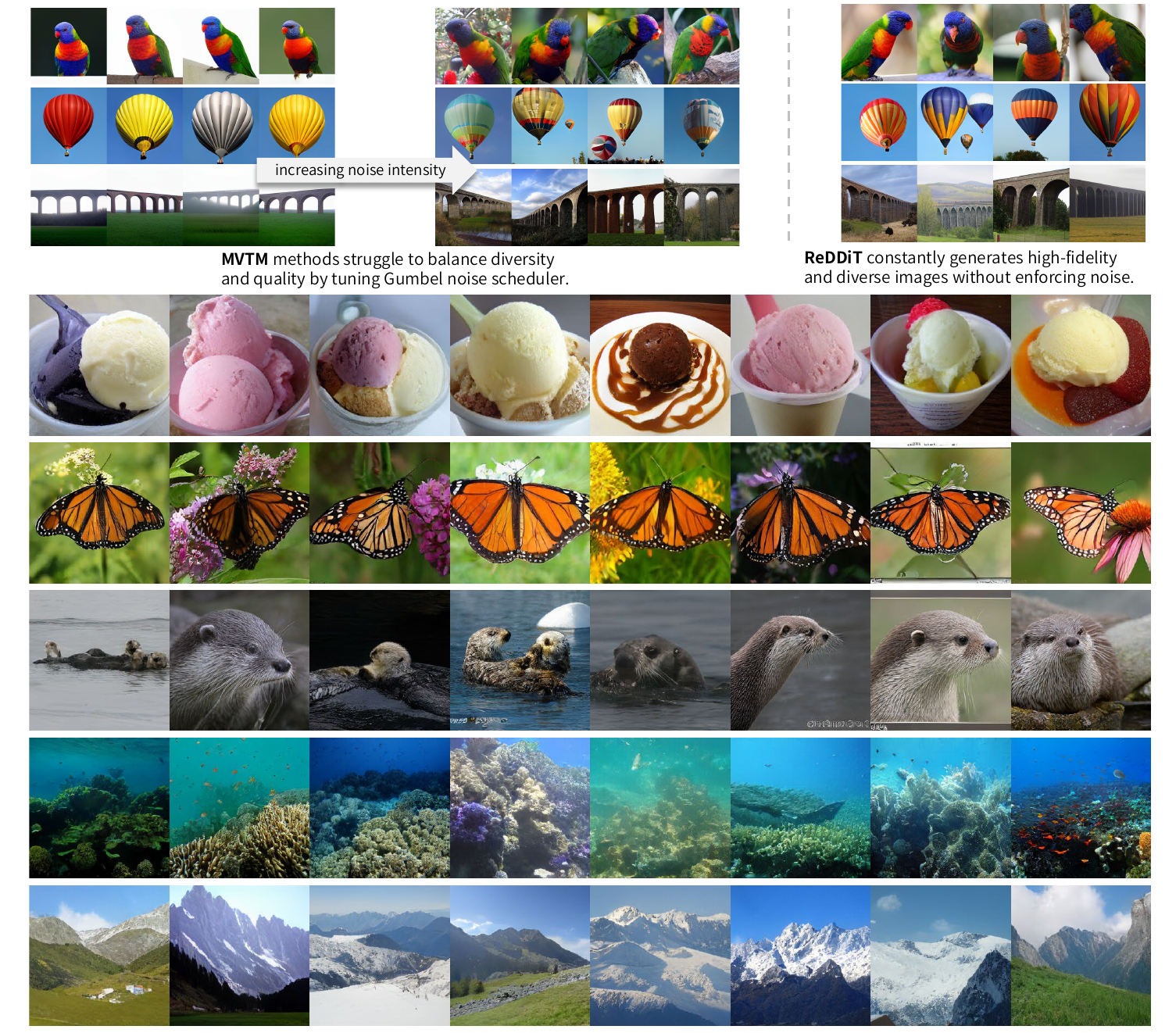}
\caption{Upper: Comparison between MVTM and our method on generated images. Below: Class-conditional generation samples of ReDDiT on ImageNet $256\times 256$.}
\label{fig:more}
\end{figure}

We provide a comparison between MVTM and our method on generated images, and more samples of ReDDiT's generation in Fig.~\ref{fig:more}.

\end{document}